\documentclass{elsart}

\usepackage{amsmath,amsfonts,amssymb,graphicx,multirow}
\usepackage{algorithm,algorithmic}
\usepackage[font=footnotesize]{caption}
\usepackage[font=footnotesize]{subcaption}
\usepackage{xcolor}
\usepackage{booktabs}
\usepackage{url}

\begin{document}

\begin{frontmatter}

\title{A GPU-Accelerated Hybrid Method for a Class of Multi-Depot Vehicle Routing Problems}

\author[LERIA]{Zhenyu Lei},
\ead{zhenyu.lei@etud.univ-angers.fr}
\author[LERIA]{Jin-Kao Hao\corauthref{cor}}
\corauth[cor]{Corresponding author.} 
\ead{jin-kao.hao@univ-angers.fr}
\address[LERIA]{LERIA, Universit$\acute{e}$ d'Angers, 2 Boulevard Lavoisier, 49045 Angers, France}

\maketitle

\begin{abstract}
Multi-depot vehicle routing problems (MDVRPs) are prevalent in a variety of practical applications. However, they are computationally challenging to solve due to their inherent complexity. This paper proposes an effective hybrid algorithm for a class of MDVRPs. The algorithm integrates a learning-driven, diversity-controlled route-exchange crossover and a multi-depot-supported feasible-and-infeasible search framework guided by a multi-penalty evaluation function. Two dedicated depot-related local search operators are incorporated to further strengthen the search capability in multi-depot settings. To improve computational efficiency and scalability, an enhanced version of the algorithm is developed that uses a tensor-based GPU acceleration combined with a novel multi-move update strategy. Extensive computational experiments on benchmark instances of three MDVRP variants show that the proposed algorithms are highly competitive with state-of-the-art methods, especially for large-scale instances.

\end{abstract}

\end{frontmatter}

\section{Introduction}
\label{sec:introduction}

The Multi-Depot Vehicle Routing Problem (MDVRP) \cite{tillman1969multiple} is a well-known combinatorial optimization problem that extends the classical Vehicle Routing Problem (VRP) \cite{dantzig1959truck,clarke1964scheduling} by considering multiple depots from which vehicles are dispatched. The MDVRP is computationally challenging because it can be reduced to the classical single-depot VRP, which is itself NP-hard \cite{lenstra1978complexity}.

The MDVRP has many applications in transportation, logistics, and supply chain management. Over the years, various problem variants have been developed to address different operational constraints and objectives. In this paper, we focus primarily on the classical MDVRP and the Multi-Depot Vehicle Routing Problem with Time Windows (MDVRPTW), which introduces time window constraints and is one of the most widely studied variants. We also consider the Multi-Depot Open Vehicle Routing Problem (MDOVRP) in our computational experiments. In this variant, vehicles are not required to return to their depots after serving customers.

The MDVRP and its variants can be described on a complete graph $\mathcal{G} = (\mathcal{V}, \mathcal{E})$, where the set of nodes $\mathcal{V}$ consists of $N_D$ depot nodes $\mathcal{V}_D$ and $N_C$ customer nodes $\mathcal{V}_C$. The edges $\mathcal{E}$ represent all connections between nodes. Each customer node $v_i \in \mathcal{V}_C$ is associated with a delivery demand $q_i$ that is served by an assigned vehicle, requiring a service time $s_i$. In the MDVRPTW, each node $v_i$ has a time window $[e_i, l_i]$ denoting the earliest and latest time for the service. For depot nodes, the time windows define the earliest departure and latest return times, but their service times are set to zero. Each edge $e_{ij} \in \mathcal{E}$ is associated with a travel distance $c_{ij}$ and a travel time $t_{ij}$, both of which are assumed to be symmetric and satisfy the triangle inequality. Each depot in $\mathcal{V}_D$ operates a fleet of up to $N_V$ homogeneous vehicles, each with a limited capacity $\mathcal{Q}$. The objective of MDVRPs is to determine a set of vehicle routes that collectively serve all customers while minimizing the total cost, typically defined as the total distance traveled by all vehicles, subject to various problem-specific constraints, as shown in Equation (\ref{eq:objective}).

\begin{align}
\label{eq:objective}
\mathrm{Minimize} \quad f(S) &= \sum_{i=1}^{M} \sum_{j=0}^{L_i} c_{n_{i,j}n_{i,j+1}}
\end{align}

In this formulation, the solution $S$ is represented as a set of routes $S = \{R_1, \dots, R_M\}$, where each route $R_i$ is a sequence of nodes $\{n_{i,0}, n_{i,1}, \dots, n_{i,L_i}, n_{i,L_i+1}\}$ visited by the $i$-th vehicle, with length $|R_i| = L_i + 2$. Here, $n_{i,0}$ and $n_{i,L_i+1}$ denote the departure and arrival depot, and $L_i$ is the number of customers served. A feasible solution requires that each route starts and ends at the same depot, i.e., $n_{i,0} = n_{i,L_i+1} \in \mathcal{V}_D$. Note that in the case of the MDOVRP, vehicles are not required to return to their depots after completing service, so the arrival depot node $n_{i,L_i+1}$ can be neglected. Additional feasibility conditions include that the number of vehicles used per depot does not exceed the available fleet size $N_V$, the total load on each vehicle does not exceed capacity $\mathcal{Q}$, and each route respects a maximum tour duration $\mathcal{D}$, defined as the total time from departure to return. For the MDVRPTW, time window constraints must be satisfied. Let $a_{n_{i,j}}$ denote the arrival time at node $n_{i,j}$. If the vehicle arrives before the earliest time $e_{n_{i,j}}$, it must wait for $w_{n_{i,j}} = \max\{e_{n_{i,j}} - a_{n_{i,j}}, 0\}$. If it arrives after the latest time $l_{n_{i,j}}$, the route is considered infeasible.

Numerous algorithms have been proposed to address the MDVRP and its variants. Section \ref{sec:literature} provides a review of relevant studies. Due to the inherent complexity of MDVRPs, no single method has been consistently proven to outperform the others across all benchmark instances. Additionally, most research has focused on small and medium-sized instances, so effective solutions for large-scale scenarios remain limited. Moreover, existing algorithms have primarily been designed for traditional CPU architectures. However, recent advancements in computer hardware have made Graphics Processing Units (GPUs) a promising alternative for accelerating computationally intensive combinatorial optimization tasks, especially for large-scale problems.

In light of these considerations and to address the underlying challenges, this work introduces MDFIHA, a novel hybrid algorithm that combines the exploration capabilities of population-based search with the exploitation strengths of local search. Its key components include: (i) a multi-depot feasible-and-infeasible search guided by a multi-penalty evaluation function and enriched with two depot-specific operators; (ii) a diversity-controlled route-exchange crossover that incorporates a learning mechanism to dynamically adjust the diversity of offspring solutions; and (iii) a fitness-and-distance-based population management strategy to maintain population health. To further address the computational challenges posed by large-scale instances, we propose an enhanced variant, MDFIHA-ETGA, which features two main improvements. First, this algorithm incorporates an efficient edge-implemented tensor-based GPU acceleration (ETGA) \cite{lei2025speedinglocaloptimizationvehicle}, which accelerates neighborhood evaluations during local search by exploiting GPU parallelism. Second, the algorithm uses a leader-follower multi-move update strategy that applies multiple moves within a single local search iteration. This strategy leverages the parallel evaluation capabilities of GPUs more fully.

We evaluate both the CPU-based MDFIHA and the GPU-enhanced MDFIHA-ETGA algorithms using widely tested benchmark instances of three well-known multi-depot routing problems: the MDVRP, the MDVRPTW, and the MDOVRP. The results demonstrate the proposed algorithms' effectiveness and robustness across various problem variants, achieving numerous improved upper bounds on the benchmark instances: 7 new bounds for the 33 MDVRP instances, 4 and 19 new bounds for the 20 classical and 28 large-scale MDVRPTW instances, respectively, and 6 new bounds for the 24 MDOVRP instances. The GPU-enhanced MDFIHA-ETGA algorithm further improves the performance of the CPU-based MDFIHA algorithm on the 28 large-scale MDVRPTW instances, achieving improved upper bounds for 26 instances and elevating 14 of the 19 new bounds from MDFIHA.

The remainder of this paper is organized as follows. Section \ref{sec:literature} presents a literature review focused primarily on the MDVRP and MDVRPTW. Section \ref{sec:algorithm} describes the proposed algorithm in detail. Section \ref{sec:experiments} reports computational results on benchmark instances to demonstrate the effectiveness of the method. Section \ref{sec:analysis} provides an in-depth analysis of the key algorithmic components. Finally, Section \ref{sec:conclusion} offers concluding remarks and outlines potential directions for future research.

\section{Literature review}
\label{sec:literature}

The Vehicle Routing Problem is a fundamental combinatorial optimization problem that has been widely studied in the literature. It was first introduced by Dantzig and Ramser \cite{dantzig1959truck} as the Truck Dispatching Problem and then was formalized as a linear optimization problem by Clarke and Wright \cite{clarke1964scheduling}. Of its many variants, the multi-depot variant (MDVRP) \cite{tillman1969multiple} stands out for its practical relevance, which better reflects real-world logistics. Because of its importance, extensive research has produced a wide range of solution methods, belonging to exact algorithms, heuristics, and metaheuristics, as surveyed in \cite{montoya2015literature,sharma2020heuristics}. This section reviews key contributions to the MDVRP and its time-constrained variant, the MDVRPTW, which is another widely studied problem, due to its practical applicability in logistics.

Laporte \cite{laporte1984optimal} proposed the first integer linear programming model for the MDVRP and developed a branch-and-bound exact algorithm. Other representative exact methods for the MDVRP were reported in \cite{laporte1988solving,baldacci2009unified,contardo2014new}. However, due to the problem's NP-hard nature, exact algorithms become impractical for large-scale instances. For instance, Contardo and Martinelli \cite{contardo2014new} reported that their exact method struggled with MDVRP instances with over 200 customers and could not solve those with over 300 customers in a reasonable amount of time. Consequently, heuristic and metaheuristic methods have become more common for solving the MDVRP.

The ``cluster-first, route-second" heuristic strategy is widely used in MDVRP research. This approach first assigns customers to depots, and then solves the routing and scheduling problems for each depot. Giosa et al. \cite{giosa2002new} focused on the assignment phase by integrating several heuristics with a modified Clarke and Wright heuristic to address the MDVRPTW. Tansini and Viera \cite{tansini2006new} introduced new proximity measures accounting for time windows and distances. More recently, Torres-P\'erez et al. \cite{torres2022new} proposed two new assignment heuristics showing competitive performance. Lim and Wang \cite{lim2005multi} compared two MDVRP approaches: a two-stage method based on the ``cluster-first, route-second" strategy, and a one-stage method that integrates assignment and routing simultaneously. They concluded that the one-stage algorithm is more effective.

Among the heuristic and metaheuristic methods for solving the MDVRP, tabu search stands out as one of the most popular methods. Renaud et al. \cite{renaud1996tabu} proposed a tabu search using the Improved Petal heuristic \cite{renaud1996improved} to generate initial solutions. Cordeau et al. \cite{cordeau1997tabu} later developed a unified tabu search heuristic applicable to multiple VRP variants, including the MDVRP. They evaluated their method using 33 benchmark instances and reported competitive results. This benchmark set remains the most widely used for the MDVRP.

Cordeau et al. \cite{cordeau2001unified} extended their approach to include time window constraints and proposed a tabu search algorithm for VRP variants with time windows, including the MDVRPTW. They introduced 20 new benchmark instances for the MDVRPTW and showed that their algorithm produced competitive results. These 20 instances remain a standard benchmark for evaluating MDVRPTW algorithms. A later refinement with forward time slack strategy further improved performance \cite{cordeau2004improved}. Finally, a parallel iterated tabu search \cite{cordeau2012parallel} produced new best-known solutions with a simple and efficient implementation.

Another notable metaheuristic for MDVRPs is Variable Neighborhood Search. Polacek et al. \cite{polacek2004variable} proposed the first VNS-based algorithm for the MDVRPTW, achieving competitive performance against existing tabu search methods. Later, they incorporated parallel cooperative schemes \cite{polacek2008cooperative}, producing new best-known solutions. Sadati et al \cite{sadati2021efficient} presented a competitive VNS-based algorithm to solve a class of multi-depot vehicle routing problems, including the MDVRP, MDVRPTW, and MDOVRP. To enhance diversification and prevent premature convergence, the algorithm incorporates a tabu-based shaking mechanism during the diversification phase. The results on standard benchmark instances demonstrate the algorithm's effectiveness and competitiveness in solving the MDVRP and its multiple variants. 

Adaptive Large Neighborhood Search has also proven to be another effective approach. In particular, Pisinger and Ropke \cite{pisinger2007general} developed a generic ALNS algorithm to solve five VRP variants, including the MDVRP. ALNS dynamically selects among a set of insertion and removal operators based on their past performance, allowing the algorithm to adaptively explore the solution space. The algorithm was tested on various benchmark instances, demonstrating strong performance and proved competitive with state-of-the-art methods.

Hybrid genetic algorithms have also been widely applied in MDVRP research. Vidal et al. \cite{vidal2012hybrid} proposed a Hybrid Genetic Search combining powerful neighborhood-based local search with effective population diversity control. This approach achieved new best-known solutions on benchmark instances of multiple VRP variants.  Additionally, Vidal et al. \cite{vidal2013hybrid} introduced an enhanced algorithm called Hybrid Genetic Search with Advanced Diversity Control for solving a broad class of time-constrained VRPs, including the MDVRPTW. Improved move evaluation strategies and robust diversity management contributed to its strong performance on existing benchmarks and on a set of new large-scale MDVRPTW instances proposed by the authors. Both studies \cite{vidal2012hybrid,vidal2013hybrid} adopted the giant-tour approach \cite{prins2004simple}. This approach first transforms a multiple routing solution into a giant-tour by concatenating all routes without delimiters, and then performs the simple order crossover, followed by an efficient \emph{Split} algorithm to obtain the offspring routing solution.

Some studies have focused on improving algorithm efficiency through neighborhood reduction. Escobar et al. \cite{escobar2014hybrid} proposed a Hybrid Granular Tabu Search using five granular neighborhoods and multiple diversification strategies to balance solution quality and computational efficiency. Experiments on benchmark instances demonstrate that the proposed method is both competitive and effective compared to state-of-the-art algorithms. Another notable study is by Tu et al. \cite{tu2014bi}, who introduced a bi-level Voronoi diagram metaheuristic to efficiently reduce the search space for solving large-scale real-world MDVRPs (up to 20,000 customers), though it was less competitive on standard benchmarks.

Despite these advancements, no single method dominates across all benchmarks. In terms of algorithmic design, most approaches follow the ``cluster-first, route-second" strategy or treat depot assignment and routing separately, rather than integrating them into a unified framework. Several local search based methods \cite{vidal2013hybrid,escobar2014hybrid,sadati2021efficient} incorporate penalty terms in the objective function to guide the search process. However, these typically neglect explicit modeling of depot-related constraints. In addition, while some efforts have aimed to improve computational efficiency, such as through parallel computing \cite{cordeau2012parallel} or neighborhood reduction techniques \cite{escobar2014hybrid,tu2014bi}, these approaches are generally implemented on traditional CPU architectures. Moreover, most crossover strategies in the MDVRP literature are based on the giant-tour representation combined with the \emph{Split} algorithm \cite{vidal2012hybrid,vidal2013hybrid}, which places strong dependence on an efficient split procedure for offspring generation. Furthermore, only a few studies have evaluated their methods on large-scale problem instances. Motivated by these observations, we propose a novel algorithm, which is described in the following section.

\section{Multi-depot feasible-and-infeasible hybrid algorithm}
\label{sec:algorithm}

The proposed MDFIHA algorithm for solving MDVRPs relies on the memetic search framework \cite{moscato2010modern}, which combines genetic search and local optimization. Specifically, MDFIHA incorporates a powerful multi-depot-supported feasible-and-infeasible search (MDFIS) along with a learning-driven diversity-controlled route-exchange crossover (DCREX) to balance intensification and diversification. To effectively handle the complex constraints and multi-depot characteristics of these problems, we design a multi-penalty evaluation function that guides the search process through feasible and infeasible regions and introduce two dedicated depot-related local search operators to enhance the search capability in multi-depot settings. To further improve computational efficiency for large-scale instances, we propose an enhanced version of the algorithm, referred to as MDFIHA-ETGA, which incorporates the edge-implemented tensor-based GPU acceleration \cite{lei2025speedinglocaloptimizationvehicle} and introduces a leader-follower multi-move update strategy to fully exploit GPU parallelism.
\begin{algorithm}[!ht] 
\caption{Main framework of MDFIHA}
\label{algo:framework} 
\begin{algorithmic}[1]
    \STATE \textbf{Input}: Instance $\mathcal{I}$, Population size $\mu$, Maximum number of generations $\varphi_{max}$, Stagnation patience threshold $\rho$, Maximum running time $\tau$.
    \STATE \textbf{Output}: The best solution $S^*$.
    \STATE $\varphi \gets 0, \varphi_{st} \gets 0$ \hfill /* Current generation and stagnation counter */
    \STATE $\mathcal{P} \gets Initialization(\mathcal{I}, \mu)$ \hfill /*Initialization in Section \ref{sec:init} */
    \STATE $S^* \gets BestSolution(\mathcal{P})$ \hfill /* Record current best solution */
    \WHILE {$\varphi \leq \varphi_{max}$ \textbf{and} $\varphi_{st} \leq \rho$ \textbf{and} $time() \leq \tau$}
        \STATE $S' \gets DCREX(\mathcal{P})$ \hfill /* diversity-controlled route-exchange crossover in Section \ref{sec:crossover} */
        \STATE $S' \gets MDFIS(S')$ \hfill /* Multi-depot-supported feasible-and-infeasible search in Section \ref{sec:search} */
        \STATE $\mathcal{P} \gets UpdatePopulation(S', \mathcal{P})$ \hfill /*Population management in  Section \ref{sec:population} */
        \IF {$f(S') < f(S^*)$}
            \STATE $S^* \gets S'$ \hfill /* Update the best solution */
            \STATE $\varphi_{st} \gets 0$
        \ELSE
            \STATE $\varphi_{st} \gets \varphi_{st} + 1$ \hfill /* Stagnation counter is incremented */
        \ENDIF
        \STATE $\varphi \gets \varphi + 1$ 
    \ENDWHILE
    \STATE \textbf{return} $S^*$ 
\end{algorithmic}
\end{algorithm}

Algorithm \ref{algo:framework} presents the main framework of the proposed algorithm. Given an instance $\mathcal{I}$, the algorithm first initializes a population of $\mu$ solutions and records the best solution $S^*$ (lines 4-5). Then in each generation, an offspring solution $S'$ is generated via DCREX (line 7), and refined using the MDFIS procedure (line 8). The improved offspring is then added to the population using a dedicated population update strategy (line 9). According to whether $S'$ improves on $S^*$, the best solution and stagnation counter are updated (lines 10-15). This process repeats until one of the stopping conditions is met (line 6): the maximum number of generations $\varphi_{\text{max}}$, the maximum patience threshold $\rho$ for stagnation generations $\varphi_{st}$, or the maximum runtime $\tau$. The best solution $S^*$ is then returned (line 18).

\subsection{Initialization}
\label{sec:init}

The initialization phase generates the starting population for the algorithm. Given the complexity of MDVRPs, we adopt a ``cluster-first, route-second" strategy and introduce a three-step initialization process: (i) depot assignment, (ii) route construction, and (iii) route pooling. In the first step, customers are assigned to their nearest depot based on travel distance. Next, in the route construction step, routes are built for each depot using a greedy insertion heuristic that considers travel distance, duration, and vehicle load, while promoting diversity. Only feasible insertions are permitted; if none are possible, a new route is created. This continues until all customers are routed or the vehicle limit $N_V$ is reached. In the final route pooling step, the constructed routes are merged into a complete solution. If any customers remain unrouted, a relaxed insertion heuristic allowing infeasible insertions is applied to ensure all customers are routed. If an initial solution is infeasible, it is still accepted and handled by subsequent procedures. 

\subsection{Multi-depot-supported feasible-and-infeasible search}
\label{sec:search}

The Multi-Depot-supported Feasible-and-Infeasible Search (MDFIS) is a core local search component of our proposed algorithm for search intensification (see Algorithm \ref{algo:mdfis}). MDFIS iteratively improves the input solution using a set of common VRP local search operators and two dedicated depot operators (line 5, Section \ref{sec:operators}), stopping when no improvement is found or a search depth threshold $\delta$ is reached (line 4). As discussed in \cite{GloverHao2011}, controlled exploration of infeasible regions can help escape local optima in complex constrained search spaces. To support this, we employ a multi-penalty evaluation function that guides the search across the neighborhoods defined by different move operators (lines 6 and 7, Section \ref{sec:evaluation}). 

Finally, to address the computational challenges of large-scale instances, we adopt the edge-implemented tensor-based GPU acceleration (ETGA) framework \cite{lei2025speedinglocaloptimizationvehicle} to enhance MDFIS (line 6, Section \ref{sec:etga}). ETGA leverages GPU parallelism to efficiently evaluate all neighboring solutions within the reduced neighborhood simultaneously and reduce computational time significantly on large instances. To further improve neighborhood exploration under ETGA, we additionally introduce a leader-follower multi-move update strategy (Section \ref{sec:multi_move}) that applies multiple independent moves within a single local search iteration. When ETGA is enabled, the move $\zeta$ (lines 6 and 8) represents a set of independent moves selected by this strategy, and additional processing steps are required (lines 3 and 10).
\begin{algorithm}[!ht] 
    \caption{Multi-depot-supported feasible-and-infeasible search}
    \label{algo:mdfis} 
    \begin{algorithmic}[1]
        \STATE \textbf{Input}: Input solution $S$.
        \STATE \textbf{Output}: Improved solution $S$.
        \STATE \textbf{if} ETGA is enabled, \textbf{then} initialize tensor $\mathbf{T}_s$ on GPU based on the solution $S$. 
        \WHILE{improvement is found \textbf{and} search depth threshold $\delta$ is not reached}
            \FOR {each operator $O_i$ in a random permutation of the operator set $\mathcal{O} = \{O_1, O_2, \dots, O_n\}$}
            \STATE \textbf{if} ETGA is enabled, \textbf{then} $\zeta \gets Evaluate(\mathbf{T}_s, S, O_i)$ \hfill /* ETGA-based neighborhood evaluation */
            \STATE \textbf{else} $\zeta \gets Evaluate(S, O_i)$ \hfill /* Traditional CPU-based neighborhood evaluation */
            \IF {The move $\zeta$ is accepted}
                \STATE Update current solution $S$ based on the move $\zeta$.
                \STATE \textbf{if} ETGA is enabled, \textbf{then} update tensor $\mathbf{T}_s$ based on current solution $S$.
            \ENDIF
            \ENDFOR
        \ENDWHILE
        \STATE \textbf{return} $S$ 
    \end{algorithmic}
\end{algorithm}

\subsubsection{Multi-penalty evaluation function}
\label{sec:evaluation}

The feasible and infeasible search dynamically explores both feasible and infeasible regions using an evaluation function with penalty terms. While prior works (e.g., \cite{vidal2013hybrid,escobar2014hybrid,sadati2021efficient}) applied similar strategies for MDVRPs, they typically address only vehicle capacity, route duration, and time window violations. To better handle the diverse constraints of MDVRPs, including multi-depot setting, we propose a unified multi-penalty evaluation function defined by Equation (\ref{eq:eval_func}) that measures the quality of a given solution $S$ during the search process.
\begin{align} 
    \label{eq:eval_func}
    \mathcal{F}(S) &= D(S) + \lambda_1 \cdot V_{1}(S) + \lambda_2 \cdot V_{2}(S) + \lambda_3 \cdot V_{3}(S) + \lambda_4 \cdot V_{4}(S)
\end{align}
Our evaluation function, which is to be minimized, takes into account not only the total travel distance $D(S)$ (the objective function), but also four normalized penalty terms for time window $V_{1}(S)$, vehicle capacity $V_{2}(S)$, route duration $V_{3}(S)$, and depot-related violations $V_{4}(S)$. It is flexible and can be adapted to different MDVRP variants by enabling or disabling specific terms (e.g., omitting the time window term for the basic MDVRP). Notably, the depot-related penalty $V_4(S)$ accounts for both route closure violations, penalizing the routes that do not start and end at the same depot, and depot capacity violations for exceeding available vehicles. The penalty coefficients $\lambda_1-\lambda_4$ are updated dynamically after each accepted move based on constraint feasibility. Each penalty coefficient is adjusted independently. If the associated constraint is violated, the coefficient is divided by an adjustment factor $\kappa \in (0, 1)$ (i.e., $\lambda_i \leftarrow \frac{\lambda_i}{\kappa}$) to prioritize satisfying that constraint; otherwise, it is multiplied by $\kappa$ (i.e., $\lambda_i \leftarrow \lambda_i \cdot \kappa$) to encourage exploration in infeasible regions. The details of the evaluation function and penalty terms are provided in the supplementary material.

\subsubsection{Move operators and evaluation}
\label{sec:operators}

MDFIS performs route optimization with four local search operators commonly used in vehicle routing (\emph{Relocate}, \emph{Swap}, \emph{2-opt*}, and \emph{2-opt}) and two proposed dedicated operators for the multi-depot setting (\emph{Depot-Insert} and \emph{Depot-Replace}). To efficiently evaluate move operators, we employ a sequence concatenation-based move evaluation \cite{lei2025speedinglocaloptimizationvehicle}, enabling constant-time computation of complex constraint violations. This method treats moves as rearrangements of route subsequences and merges them using a concatenation operator $\oplus$.

Among the four common routing operators, the \emph{Relocate} operator moves a node sequence within the same route or to another route. The \emph{Swap} operator exchanges two node sequences between or within routes. In our implementation, the length of each node sequence for both \emph{Relocate} and \emph{Swap} is limited to 1 or 2. For symmetric problems like MDVRP, 2-node sequences can also be reversed to enhance search capacity. The \emph{2-opt*} operator works on two routes by removing one edge from each route and exchanging the resulting subsequences to create new routes. The \emph{2-opt} operator modifies a single route by removing two non-adjacent edges and reconnecting the subsequences in reverse order. Note that \emph{2-opt} is not applied in MDVRPTW instances, as the presence of time window constraints introduces asymmetry.
\begin{figure}[!htbp]
    \centering
    \begin{subfigure}[t]{0.6\textwidth}
        \includegraphics[trim=10 10 10 10,clip,width=\textwidth]{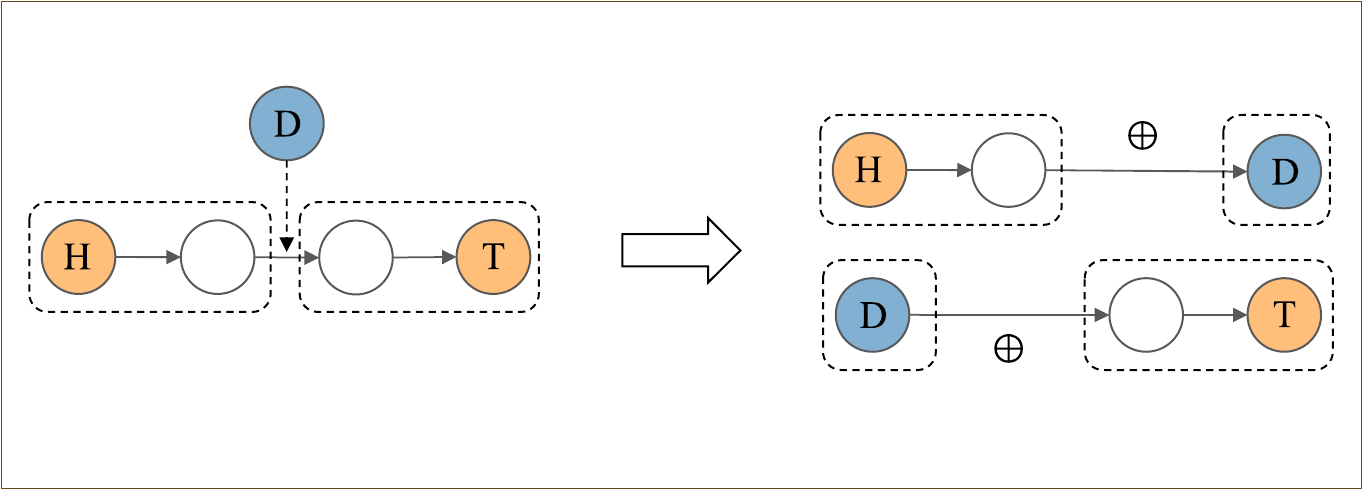}
        \caption{\emph{Depot-Insert}}
        \label{fig:depot-insert}
    \end{subfigure}
    \hfill
    \begin{subfigure}[t]{0.6\textwidth}
        \includegraphics[trim=10 10 10 10,clip,width=\textwidth]{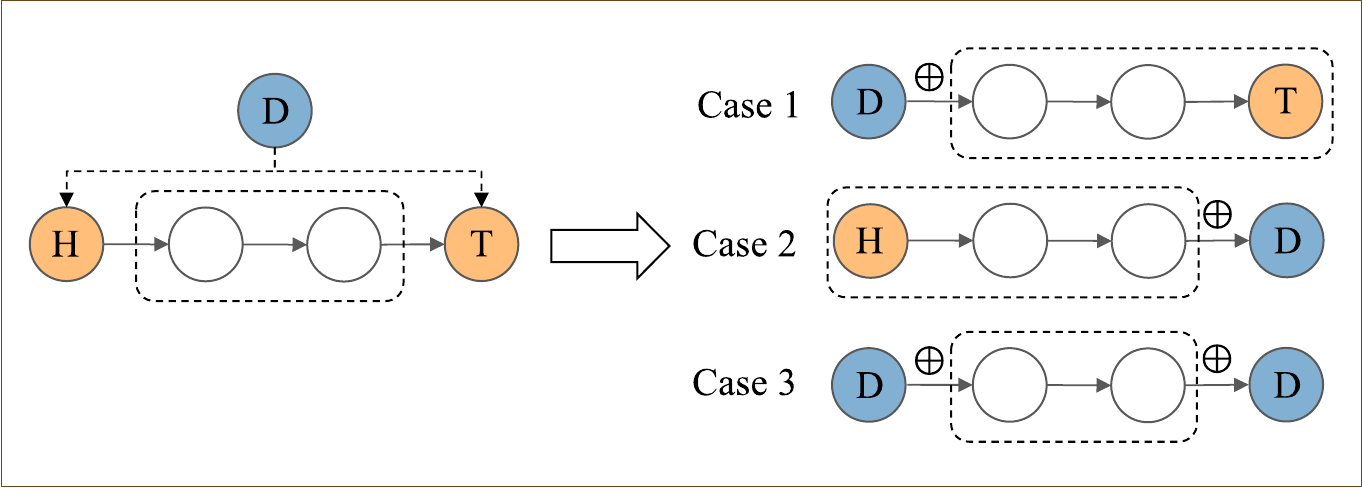}
        \caption{\emph{Depot-Replace}}
        \label{fig:depot-replace}
    \end{subfigure}
    \caption{Illustration of the two dedicated depot operators \emph{Depot-Insert} and \emph{Depot-Replace}, along with their concatenation operations. Yellow nodes represent the original depot nodes, while ``H'' and ``T'' denote the head and tail of the route, respectively. The blue ``D'' nodes represent the new depot being inserted or used for replacement.}
\label{fig:depot-operators}
\end{figure}

To enhance performance in the multi-depot setting, we design two dedicated depot operators: \emph{Depot-Insert} and \emph{Depot-Replace} (see Figure \ref{fig:depot-operators}). They build on the depot-related penalty term $V_4(S)$ from Equation (\ref{eq:eval_func}), which allows for routes to have different departure and arrival depots. The \emph{Depot-Insert} operator introduces a depot node into a route, splitting the original route into two separate routes. The \emph{Depot-Replace} operator replaces the depot of a route with a new route by evaluating three cases: replacing the departure depot, replacing the arrival depot, or replacing both. Among them, the best resulting configuration is applied. The time complexity is $O(N_D \cdot (M + N_C))$ for \emph{Depot-Insert} and $O(N_D \cdot M)$ for \emph{Depot-Replace}, where $N_D$ is the number of depots, $M$ is the number of routes, and $N_C$ is the number of customers.

\subsubsection{Neighborhood reduction}
\label{sec:pruning}

We adopt the best-improvement strategy for local search, which requires exploring the entire neighborhood of each operator to identify the best neighboring solution. However, exhaustively searching the neighborhood is computationally intensive, and becomes a bottleneck for large instances. To address this issue, we apply a neighborhood reduction technique to eliminate unpromising neighboring solutions.

We use two pruning strategies to reduce the neighborhood. First, based on the time window constraints, any edge $e_{ij} \in \mathcal{E}$ in the complete graph $\mathcal{G}=(\mathcal{V}, \mathcal{E})$ is removed if visiting $v_j$ after $v_i$ violates the latest allowed time, i.e., $e_i + s_i + t_{ij} > l_j$. Second, the neighborhood exploration is restricted to the $\theta$ nearest neighbors of each node. The parameter $\theta$ controls the neighborhood granularity, retaining only the $\theta$ most correlated neighbors. The neighbors are selected according to the correlation metric $\eta_{ij}$ in Equation (\ref{eq:correlation}) inspired by \cite{vidal2013hybrid,liu2021memetic}, which accounts for both spatial proximity and time-window compatibility.

\begin{align}
\label{eq:correlation}
\eta_{ij}
&= c_{ij} + \Gamma \Big(
\alpha \cdot \max\{e_j - l_i - s_i - t_{ij}, 0\} \\
&+ \beta \cdot \max\{l_i + s_i + t_{ij} - l_j, 0\}
\Big), \nonumber \\
\mathrm{where} \quad
\Gamma
&= \frac{\sum_{e_{uv}\in \mathcal{E}} c_{uv}}
{\sum_{e_{uv}\in \mathcal{E}} t_{uv}}, \qquad v_i,v_j \in \mathcal{V}. \nonumber
\end{align}

The factor $\Gamma$ represents the average distance-time ratio over all edges and is used to scale time-related penalties into distance units. The constants $\alpha$ and $\beta$ control the relative importance of early-arrival and late-arrival incompatibility, respectively. In this work, we adopt $\alpha=1$ and $\beta=10$ following \cite{liu2021memetic} to preserve the original scale of early-arrival penalties while emphasizing the greater severity of late arrivals.

\subsubsection{Edge-implemented TGA}
\label{sec:etga}

For large-scale instances, we adopt the edge-implemented tensor-based GPU acceleration \cite{lei2025speedinglocaloptimizationvehicle} to accelerate move evaluation within the reduced neighborhood by leveraging the parallel computing capabilities of GPUs.

The tensor-based GPU acceleration (TGA) framework proposed in \cite{lei2025speedinglocaloptimizationvehicle} speeds up local search for vehicle routing problems, by accelerating the move evaluation for commonly used local search operators, \emph{Relocate}, \emph{Swap}, \emph{2-opt*}, and \emph{2-opt}. This is achieved by tensorizing the move evaluation based on an attribute-based solution representation and exploiting massive GPU parallelism, leading to substantial computational gains over traditional CPU-based implementations of local search operators.

Specifically, the attribute-based solution representation is based on a four-dimensional tensor $\mathbf{T}_s \in \mathbb{R}^{I \times J \times K \times K}$ stored on the GPU, where $I$ denotes the number of maintained attributes, $J$ the maximum number of routes, and $K$ the maximum route length. By extracting and concatenating relevant attribute tensors from the solution tensor $\mathbf{T}_s$, move evaluation can be performed through a sequence of highly parallel tensor operations on the GPU.

Building upon this framework, Lei et al. \cite{lei2025speedinglocaloptimizationvehicle} introduced the edge-implemented TGA to support neighborhood reduction and further decrease the computational cost of move evaluation. ETGA employs a binary mask tensor $\mathbf{M} \in \{0,1\}^{|\mathcal{V}| \times |\mathcal{V}|}$ to encode the result of a predefined neighborhood reduction strategy. Each element $\mathbf{M}_{ij}$ indicates whether a valid edge exists between nodes $v_i$ and $v_j$. By using this mask, ETGA restricts tensor extraction and move evaluation to valid node pairs only, thereby avoiding unnecessary computations over the full neighborhood and significantly improving efficiency.

In this work, we adopt ETGA to implement the neighborhood reduction strategy described in Section~\ref{sec:pruning} to further improve the computational efficiency of the local search, particularly for large-scale instances. Owing to the flexibility and extensibility of its attribute-based solution tensor representation, ETGA can be readily adapted to support the proposed multi-penalty evaluation function for MDVRPs by incorporating the required attributes. Note that ETGA is applied only to the four local operators (\emph{Relocate}, \emph{Swap}, \emph{2-opt*}, and \emph{2-opt}); the two proposed depot-specific operators (\emph{Depot-Insert} and \emph{Depot-Replace}) remain CPU-based due to their low computational complexity.

\subsubsection{Leader-follower multi-move update strategy}
\label{sec:multi_move}

As discussed in Section~\ref{sec:etga}, ETGA exploits GPU parallelism to evaluate all candidate moves within a reduced neighborhood simultaneously. However, the original implementation of ETGA \cite{lei2025speedinglocaloptimizationvehicle} updates the current solution by applying only the single best move in each local search iteration, similar to the conventional CPU-based local search. Consequently, a significant portion of the evaluated moves go unused, and the available parallelism remains underutilized.

To better leverage GPU parallelism, several studies on GPU-based local search for the TSP \cite{qiao2018massive,qiao2020multiple,robinson2018analysis} have applied multiple moves per iteration. Motivated by these studies, we introduce a multi-move update strategy within the ETGA framework to further enhance the efficiency of the proposed local search component MDFIS. However, unlike the TSP considered in \cite{qiao2018massive,qiao2020multiple,robinson2018analysis}, MDVRPs involve more complex constraints, while MDFIS explores both feasible and infeasible regions guided by a multi-penalty evaluation function, making it difficult to directly adopt existing multi-move strategies.

Therefore, we propose a leader-follower multi-move update strategy specifically tailored to the proposed MDFIS under the ETGA framework. The \emph{leader} move is defined as the single best move identified according to the evaluation function, which is consistent with the traditional single-move local search. The \emph{follower} moves are selected from the remaining candidates that satisfy two conditions: (i) a strict decrease in travel distance (i.e., $\Delta D < 0$), and (ii) no variation in any constraint-related penalty term (i.e., $\Delta V_i = 0$, $i = 1,2,3,4$). Identifying the leader and follower moves can be efficiently performed on the GPU during move evaluation. Then, all candidate moves are transferred to the CPU for selection. 

The move selection process proceeds as follows. The leader move is always applied to maintain consistency with single-move local search behavior. The follower moves are sorted in ascending order of their evaluation scores and selected greedily, provided that they do not conflict with any previously selected moves. Two moves are considered conflicting if they involve the same route. Through this process, a set of independent moves is identified and applied simultaneously to update the current solution in each local search iteration.

By evaluating and applying multiple independent moves within a single iteration, the proposed multi-move update strategy more effectively exploits GPU parallelism under the ETGA framework and enhances overall search efficiency, particularly for large-scale instances. Moreover, retaining the best move as the leader while restricting follower moves to those that improve travel distance without increasing constraint violations ensures that the search process remains well guided by the multi-penalty evaluation function, even when multiple moves are applied concurrently. 

\subsection{Diversity-controlled route-exchange crossover}
\label{sec:crossover}

Unlike giant-tour-based crossover approaches \cite{prins2004simple,vidal2012hybrid,vidal2013hybrid}, which require a carefully designed and efficient split procedure, or edge-assembly-crossover-based methods \cite{nagata1997edge,he2023general,he2023memetic}, which face challenges in handling complex multi-depot problems, such as MDVRPTW, where time windows introduce asymmetry, this work adopts a route-based crossover paradigm that provides more flexibility for multi-depot settings and complex constraints.

The proposed Diversity-Controlled Route-Exchange Crossover (DCREX) generates promising offspring solutions from multiple parent solutions through route exchanges guided by a learning-driven diversity control mechanism. DCREX is inspired by the Adaptive Route Inheritance Crossover (ARIX) introduced in \cite{lei2024memetic}, which was originally developed for the vehicle routing problem with simultaneous pickup and delivery and time windows (VRPSPDTW). Although ARIX incorporates strategies tailored to the characteristics of VRPSPDTW, its underlying principles, including route-based multi-parent crossover and learning-based decision mechanisms, are broadly applicable across different VRP variants.

Building on these general ideas, we design DCREX as a generalized route exchange crossover with adaptive diversity control, which also supports multi-depot settings in MDVRPs. The objective is to achieve a better balance between solution quality and population diversity during the evolutionary search process. Specifically, DCREX generates an offspring solution by selecting one parent as the main parent and selectively replacing a subset of its routes with routes inherited from other parent solutions, with the route exchange process guided by a learning-based diversity control mechanism.

\subsubsection{General crossover procedure}
\label{sec:general}
\begin{algorithm}[!ht] 
\caption{Diversity-controlled route-exchange crossover}
\label{algo:dcrex} 
\begin{algorithmic}[1]
    \STATE \textbf{Input}: Population $\mathcal{P}$.
    \STATE \textbf{Output}: Offspring solution $S'$.
    \STATE Let $P_m$ be the main parent randomly selected from $\mathcal{P}$ \hfill /* Main parent */
    \STATE $S' \gets P_m$ \hfill /* Initialize offspring using main parent */
    \STATE Determine diversity degree $\sigma_i = \frac{i}{20}, i \in \{0, 1 \cdots, 19\}$ based on the learning mechanism \hfill /* Section \ref{sec:learning} */
    \STATE $\sigma_{min} = \sigma_i \cdot (N_C + M_{P_m})$ \hfill /*  Calculate the diversity bounds $\sigma_{min}$ and $\sigma_{max}$ based on number of
    \STATE $\sigma_{max} = \sigma_{i+1} \cdot (N_C + M_{P_m})$ \hfill customer nodes $N_C$ and number of routes $M_{P_m}$ in main parent $P_m$*/
    \STATE $\sigma_{cur} = 0$  \hfill /* Initialize current diversity value */
    \FOR{$P \in \mathcal{P} \setminus \{P_m\}$}
        \STATE $Score_{ij} = - (E^I_{ij} - N^R_{ij} - N^M_{ij} - 10 \cdot N^C_{ij}) $ \hfill /* Calculate the score of each route pair $\{R_i, R_j\}, R_i \in P_m, R_j \in P$ */
        \STATE Randomly select $\{R_i^*, R_j^*\}$ from the 5 lowest-scored route pairs \hfill
        \STATE $S' \gets (S' \setminus R_i^*) \cup R_j^*$ \hfill /* Remove the route $R_i^*$ and add the route $R_j^*$ */
        \STATE $\Delta \sigma = E^I_{ij} + N^R_{ij} + N^M_{ij}$ \hfill /* Calculate the variation of diversity */
        \STATE $\sigma_{cur} = \sigma_{cur} + \Delta \sigma$ \hfill /* Update the current diversity value */
        \STATE \textbf{if} $\sigma_{cur} \ge \sigma_{max}$ \textbf{then} Continue to next parent \hfill
        \STATE \textbf{else if} $\sigma_{cur} \ge \sigma_{min}$ \textbf{then} Break the loop \hfill
    \ENDFOR
    \STATE Remove redundant nodes in $S'$ \hfill
    \STATE $U \gets \{n \in \mathcal{V} | n \notin S'\}$ \hfill /* Unrouted nodes */
    \STATE Insert the unrouted nodes in $U$ into $S'$ using the insertion operator $\boxdot$ \hfill
    \STATE \textbf{return} $S'$ 
\end{algorithmic}
\end{algorithm}

As shown in Algorithm \ref{algo:dcrex}, the DCREX crossover uses a multi-parent design to increase solution diversity, while the number of parents is determined adaptively. A random parent is first selected as the main parent $P_m$ (line 3), and the offspring $S'$ is initialized as a copy of $P_m$ (line 4). A target diversity degree $\sigma_i$ is then determined via a learning mechanism (line 5, see next section) from 20 discrete levels. The corresponding diversity bounds $\sigma_{\min}$ and $\sigma_{\max}$ are computed (lines 6-7), and the current diversity $\sigma_{\text{cur}}$ is set to zero (line 8). DCREX then iterates over other parents to exchange routes (lines 9-17). For each candidate parent $P$, DCREX scores route pairs $\{R_i, R_j\}$, where $R_i$ is from $P_m$ and $R_j$ from $P$ (line 10). The score accounts for four factors: (i) newly introduced edges $E^I_{ij}$, (ii) missing nodes $N^R_{ij}$, (iii) redundant nodes $N^M_{ij}$ appearing more than once, and (iv) conflicting nodes $N^C_{ij}$ that already exist in the introduced routes. Each factor is weighted to reflect its relative importance, with conflicting nodes heavily penalized to avoid duplicates. Lower scores indicate better route pairs. One of the top five ranked pairs $\{R_i^*, R_j^*\}$ is selected randomly (line 11), and $R_i^*$ in $S'$ is replaced by $R_j^*$ (line 12). The resulting diversity variation $\Delta \sigma$ is computed (line 13), and $\sigma_{\text{cur}}$ is updated (line 14). If $\sigma_{\text{cur}}$ exceeds the upper bound $\sigma_{\max}$, indicating the current parent cannot provide suitable routes, the algorithm continues with the next parent (line 15). If $\sigma_{\text{cur}}$ exceeds the lower bound $\sigma_{\min}$, meaning the desired diversity degree has been reached, the loop breaks (line 16).

After the route exchange, the repair procedure \cite{lei2024memetic} is applied to ensure all customers are properly routed (lines 18-20). Specifically, redundant nodes are removed from the offspring $S'$ (line 18), and unrouted nodes are inserted using an insertion operator $\boxdot$ (lines 19-20), selected via a learning mechanism (see next section) among five operators. Feasible best insertion (FBI) and infeasible best insertion (IBI) both prioritize minimal cost increase; FBI creates a new route if no feasible insertion is possible, while IBI allows infeasible insertions without adding new routes. Feasible regret insertion (FRI) and infeasible regret insertion (IRI) evaluate each insertion based on the regret value, defined as the difference in travel cost between the best and second-best insertions. Random insertion (RI) inserts an unrouted customer randomly into a randomly selected route. If the resulting offspring is still infeasible, it is then addressed and refined by the MDFIS procedure (Section \ref{sec:search}). In the worst case, the time complexity of DCREX is $\mathcal{O}(|\mathcal{P}| \cdot M^2)$, where $|\mathcal{P}|$ denotes the number of solutions in the population and $M$ represents the average number of routes per solution. An illustration of the DCREX crossover is provided in the supplementary material.

\subsubsection{Learning-based decision mechanism}
\label{sec:learning}

During the crossover process, DCREX needs to determine the diversity degree and the insertion operator $\boxdot$ for unrouted nodes. For this purpose, it uses a learning mechanism based on the discounted UCB1 algorithm from the multi-armed bandit framework to inform these decisions. Unlike the standard UCB1 used in \cite{lei2024memetic}, the discounted variant uses a discount factor ($\gamma_d = 0.99$) to prioritize recent rewards and limit the impact of outdated data. 

\begin{align}
\label{eq:ucb1}
\text{UCB1}_i(t) &= \frac{\mathcal{R}^d_i(t)}{N_i(t)} + \sqrt{ \frac{2 \ln \left( \sum_{j=1}^{|\mathcal{A}|} N_j(t) \right)}{N_i(t)} } \\
\mathrm{where} \quad
\mathcal{R}^d_i(t) &= \sum_{\tau=1}^{t} \gamma_d^{t-\tau} \mathcal{R}_i(\tau) \cdot 1_{\{a_\tau = i\}} \nonumber \\
N_i(t) &= \sum_{\tau=1}^{t} \gamma_d^{t-\tau} \cdot 1_{\{a_\tau = i\}} \nonumber
\end{align}
For each action $a_i \in \mathcal{A}$, the discounted UCB1 algorithm computes the upper confidence bound $\text{UCB1}_i(t)$ at step $t$ (Equation (\ref{eq:ucb1})) using its discounted reward $\mathcal{R}^d_i$ and selection count $N_i$. The action with the highest value is selected at each step. In our case, the actions $\mathcal{A}$ includes 20 diversity degrees $\{0, 1, \ldots, 19\}$ and the five insertion operators presented in Section \ref{sec:general}. The reward $\mathcal{R}$ for each action is the improvement ratio of the offspring $S'$ over the main parent $P_m$, defined as $\mathcal{R} = 100 \times \frac{f(P_m) - f(S')}{f(P_m)}$.

\subsection{Population management}
\label{sec:population}

During the execution of the algorithm, the population $\mathcal{P}$ is expanded with new offspring solutions to the size of $1.5\mu$ for $\frac{\mu}{2}$ generations \cite{vidal2012hybrid}. $\mathcal{P}$ is then reduced back to its original size $\mu$ by removing the worst solutions according to the fitness-and-distance-based scoring function $\chi$ defined by Equation (\ref{eq:fitness_func}).
\begin{align}
    \label{eq:fitness_func}
    \chi(S) &= \frac{|\mathcal{P}| - \text{Rank}_f(S)}{|\mathcal{P}|} + \xi \cdot \frac{|\mathcal{P}| - \text{Rank}_{D_{\mathcal{P}}}(S)}{|\mathcal{P}|} \\
    \text{where} \quad 
    \label{eq:pop_dist}
    D_{\mathcal{P}}(S) &= \frac{1}{5} \sum_{S_i \in \mathcal{N}_5(S)} D_{S}(S, S_i), \quad S_i \neq S  \\
    \label{eq:sol_dist}
    D_{S}(S_1, S_2) &= 100 \cdot \left(1 - \frac{|\mathcal{E}_{S_1} \cap \mathcal{E}_{S_2}|}{\max\{|\mathcal{E}_{S_1}|, |\mathcal{E}_{S_2}|\}}\right), \quad \mathcal{E}_{S_1}, \mathcal{E}_{S_2} \in \mathcal{E} 
\end{align}
The scoring function $\chi$ assigns a fitness score to each solution $S$ in the population $\mathcal{P}$ by considering two factors: the rank of $S$ with respect to the objective value $\text{Rank}_f{(S)}$ and its rank based on the distance within the population $\text{Rank}_{D_{\mathcal{P}}}(S)$. The parameter $\xi$ balances the trade-off between solution quality and diversity. A low value of $\chi(S)$ indicates a solution of low fitness. 

To quantify diversity, we measure the distance between solutions using Equation (\ref{eq:sol_dist}), where $\mathcal{E}_{S_1}$ and $\mathcal{E}_{S_2}$ represent the sets of edges in solutions $S_1$ and $S_2$, respectively. The distance of a solution $S$ from the population $\mathcal{P}$ is defined as the average distance to its 5 nearest neighbors in $\mathcal{P} \setminus \{S\}$, denoted by $\mathcal{N}_5(S)$, as shown in Equation (\ref{eq:pop_dist}).

\subsection{Discussion}
\label{sec:discussion}

The proposed MDFIHA algorithm introduces several novel components to address the specific challenges of multi-depot vehicle routing problems.

First, unlike the ``cluster-first, route-second" strategy commonly adopted in existing MDVRP studies \cite{giosa2002new,tansini2006new,torres2022new}, which treats depot assignment and routing as two separate phases, MDFIHA is built upon the principle of jointly optimizing both aspects within a unified framework. This is achieved through its local search component MDFIS (Section \ref{sec:search}), based on a multi-penalty evaluation function and dedicated depot-related local search operators. Although penalty-based local search has been widely used in the literature \cite{vidal2013hybrid,escobar2014hybrid,sadati2021efficient}, existing approaches typically do not explicitly model depot-related violations. In MDFIS, the proposed multi-penalty evaluation function includes a dedicated term for depot-related violations, thereby explicitly accounting for depot assignment and route closure constraints. Furthermore, two specialized depot operators, namely \emph{Depot-Insert} and \emph{Depot-Replace}, are introduced to directly modify depot assignments during the search process. By allowing temporary route-closure inconsistencies and enabling depot-changing moves, MDFIS explores neighborhoods that simultaneously modify both routing and depot structures within a single local optimization procedure.

Second, to address the computational challenges of neighborhood exploration, particularly for large-scale instances, MDFIHA incorporates an edge-implemented tensor-based GPU acceleration technique \cite{lei2025speedinglocaloptimizationvehicle} to accelerate move evaluation. Unlike CPU-based parallel methods \cite{cordeau2012parallel}, ETGA offloads the most computationally intensive components of neighborhood evaluation to the GPU and exploits its massive parallelism to achieve substantial efficiency gains. In this work, ETGA is extended to support multi-depot settings and is integrated into the core local search operators of MDFIHA, marking the first ETGA application to these MDVRPs. This integration significantly improves the computational efficiency and enhances the scalability of the proposed algorithm when solving large-scale MDVRP instances. Furthermore, MDFIHA-ETGA introduces a novel leader-follower multi-move update strategy within the ETGA framework. This strategy maximizes GPU utilization and enhances the search capability by applying multiple independent moves simultaneously in each local search iteration.

Third, MDFIHA's crossover component DCREX (Section \ref{sec:crossover}) incorporates learning-based diversity control at the route-exchange level to help improve solution quality and maintain population diversity during the search process. The underlying route-based crossover paradigm offers greater flexibility in handling multi-depot settings and complex constraints than giant-tour-based methods \cite{prins2004simple,vidal2012hybrid,vidal2013hybrid}, which require an efficient split algorithm to decode the resulting giant-tour into a route solution after applying a simple crossover, such as ordered crossover, on a giant-tour representation. Our DCREX crossover also differs from the popular edge-assembly-crossover approaches, such as those described in \cite{nagata1997edge,he2023general,he2023memetic}, which have proven effective for the TSP and certain symmetric VRP variants. However, these approaches face difficulties when applied to complex MDVRPs, especially MDVRPTW, where time windows introduce asymmetry. Our route-based DCREX naturally accommodates depot assignments and problem-specific constraints by directly exchanging entire routes between parent solutions, making it particularly well suited to multi-depot scenarios.

Finally, DCREX is inspired by the general ideas of ARIX \cite{lei2024memetic}, including route-based multi-parent crossover and learning-based decision mechanisms. However, DCREX is designed as a more general and flexible crossover approach. Unlike ARIX, which relies on mechanisms specifically designed for the hierarchical objective structure of the VRPSPDTW, including adaptive parent selection and route inheritance control, DCREX employs a learning-based strategy to dynamically determine the target diversity level of offspring solutions. To support this process, DCREX introduces a dedicated diversity scoring mechanism and employs an adaptive multi-parent design driven by diversity bounds. As a result, DCREX enables diversity control while remaining applicable to a broader class of routing problems.

\section{Computational results}
\label{sec:experiments}

This section presents computational results for MDFIHA and MDFIHA-ETGA on standard benchmarks and compares them with state-of-the-art methods.

\subsection{Benchmark instances}
\label{sec:benchmark}

The performance of the proposed algorithm is evaluated on benchmark instances of the classical MDVRP and two variants MDVRPTW and MDOVRP to show the algorithm's robustness across multiple problems. The benchmark instances used in this study include:

\begin{itemize}
    \item \emph{C97}: 33 widely studied MDVRP instances from \cite{cordeau1997tabu} with 48 to 360 customers and 2 to 9 depots. This set includes 23 instances from the literature and 10 \emph{pr} instances generated with relaxed fleet-size limits. As some studies applied tighter limits, potentially causing result inconsistencies, we follow the original settings by default and also test the tighter version, denoted as \emph{C97-T}.
    \item \emph{C01}: 20 MDVRPTW instances from \cite{cordeau2001unified}, based on the 10 \emph{pr} instances from \cite{cordeau1997tabu} with added narrow or wide time windows. They have 48 to 288 customers and 4 to 6 depots. The last 10 instances were reported with relaxed fleet-size limits in \cite{vidal2013hybrid}, possibly due to an error. We also test these 10 instances under relaxed constraints, denoted as \emph{C01-R}.
    \item \emph{V13}: 28 large-scale MDVRPTW instances with 360 to 960 customers and 4 to 12 depots from \cite{vidal2013hybrid}. To the best of our knowledge, these instances have only been tested in the original study \cite{vidal2013hybrid}.
    \item \emph{L14}: 24 MDOVRP instances from \cite{liu2014hybrid}, derived from \emph{C97}, with 48 to 288 customers and 4 to 6 depots. Following common practice in the literature, fleet size limits $N_V$ and tour duration constraints $\mathcal{D}$ are relaxed in these instances.
\end{itemize}

\subsection{Experimental protocol and parameter tuning}
\label{sec:exp_protocol}

Both MDFIHA and its GPU-enhanced version MDFIHA-ETGA are implemented in C++. MDFIHA-ETGA leverages edge-implemented tensor-based GPU acceleration via the PyTorch C++ API. Experiments were conducted on a computer with an AMD EPYC 7282 2.8 GHz CPU. For MDFIHA-ETGA, an NVIDIA V100 GPU with 32 GB memory was used. The source code of our algorithms and our reported solutions will be made available upon paper acceptance.
\begin{table}[!htbp]
\caption{Parameter tuning results.}
\label{tab:params}
\centering
{
\resizebox{0.8\textwidth}{!}{%
\begin{tabular}{lllr}
\hline
Parameter & Description & Candidate values & \multicolumn{1}{l}{Final value} \\ \hline
$\mu$ & population size & \{10, 20, 30, 40, 50\} & 20 \\
$\theta$ & granularity threshold & \{25, 50, 75, 100\} & 50 \\
$\delta$ & local search depth & \{100, 200, 300, 400, 500\} & 500 \\
$\kappa$ & coefficient adjustment factor & (0, 1) & 0.5 \\
$\xi$ & fitness function coefficient & (0, 1) & 0.7 \\ \hline
\end{tabular}%
}
}
\end{table}

To balance solution quality and computational efficiency, several parameters were tuned empirically. The maximum number of generations $\varphi_{max}$ was set to 5000, and the stagnation patience $\rho$ was set to 500. Five critical parameters, including population size $\mu$, granularity threshold $\theta$, local search depth $\delta$, coefficient adjustment factor $\kappa$, and fitness function coefficient $\xi$, were tuned using the automatic parameter configuration tool Irace \cite{lopez2016irace}, with 10 randomly selected instances and a tuning budget of 1000. Table \ref{tab:params} summarizes the candidate ranges and best values. These settings were used consistently in all experiments. In the experiments, each instance was solved 10 times independently, except for the MDVRPTW \emph{V13} instances, which were solved 5 times in accordance with \cite{vidal2013hybrid}.

\subsection{Reference algorithms}
\label{sec:reference}

For each benchmark set, we compare our algorithm's results with the best-known solutions (BKS) and the state-of-the-art algorithms that have contributed to achieving these BKS (see Table \ref{tab:ref_algo}). Note that our MDFIHA algorithm is tested on all sets, while the GPU-enhanced MDFIHA-ETGA algorithm is only applied to the large-scale \emph{V13} instances.
\begin{table}[htbp]
\caption{Reference algorithms for benchmark sets.}
\label{tab:ref_algo}
\centering
{
\resizebox{\textwidth}{!}{%
\begin{tabular}{llll}
\hline
Reference algorithm & Approach & Problems & Benchmarks \\
\hline
CGL \cite{cordeau1997tabu} & Tabu Search & MDVRP & \emph{C97} \\
PR \cite{pisinger2007general} & Adaptive Large Neighborhood Search & MDVRP & \emph{C97} \\
VCGLR \cite{vidal2012hybrid} & Hybrid Genetic Search & MDVRP & \emph{C97} \\
ELTG \cite{escobar2014hybrid} & Hybrid Granular Tabu Search & MDVRP & \emph{C97} \\
VCGP \cite{vidal2013hybrid} & Hybrid Genetic Search & MDVRPTW & \emph{C01}, \emph{C01-R}, \emph{V13} \\
JB \cite{brandao2020memory} & Iterated Local Search & MDOVRP & \emph{L14} \\
RM \cite{lalla2021mathematical} & Exact method & MDOVRP & \emph{L14} \\
SCA \cite{sadati2021efficient} & Variable Tabu Neighborhood Search & MDVRP, MDVRPTW, MDOVRP & \emph{C97}, \emph{C97-T}, \emph{C01}, \emph{L14} \\
MDFIHA (Ours) & Memetic Algorithm & MDVRP, MDVRPTW, MDOVRP & \emph{C97}, \emph{C97-T}, \emph{C01}, \emph{C01-R}, \emph{V13}, \emph{L14} \\
MDFIHA-ETGA (Ours) & Memetic Algorithm with GPU acceleration & MDVRPTW & \emph{V13}
\\ \hline
\end{tabular}%
}
}
\end{table}

To account for hardware differences, we use CPU benchmark scores PassMark (\url{https://www.passmark.com/}) to normalize the reported computation times of the reference algorithms, using our AMD EPYC 7282 as the baseline. The processor information and conversion ratio details are provided in the supplementary material.

\subsection{Computational results and comparisons}

Table \ref{tab:summary} summarizes the comparative results of the proposed algorithm against various reference algorithms across different benchmark instance sets. It specifies the problem type, the benchmark set, and the number of instances (\emph{\#Instances}). For each algorithm, the table reports the average best objective function value (${\bar{f}_{Best}}$), the average computation time ($\bar{t}$), the normalized average computation time ($\gamma \cdot \bar{t}$) and the average gap relative to BKS ($\overline{\text{Gap}}$). The gap is calculated as $\text{Gap}(\%) = \frac{f - \text{BKS}}{\text{BKS}} \times 100$, where a negative value indicates that the algorithm has found a new improved upper bound. The number of wins (\emph{\#Wins}), ties (\emph{\#Ties}), and losses (\emph{\#Losses}) indicates cases where the MDFIHA performs better, equally, or worse than the reference algorithm, respectively. To assess statistical significance, the Wilcoxon signed-rank test was applied at a 0.05 confidence level, with the resulting \emph{p}-values also included in the table. The symbol `-' denotes unavailable results. Detailed results for all benchmark sets are provided in Tables~A.IV-A.IX in the supplementary material. 
\begin{table}[!htbp]
\caption{Summary of comparative results on benchmarks between the proposed algorithms and the reference algorithms.}
\label{tab:summary}
\centering
{
\resizebox{0.95\textwidth}{!}{%
\begin{tabular}{llllrrrrllll}
\hline
Problem & Benchmark & \#Instances & Algorithm & \multicolumn{1}{c}{\rule{0pt}{2.5ex} ${\bar{f}_{Best}}$} & \multicolumn{1}{c}{$\bar{t}$ (s)} & \multicolumn{1}{c}{$\gamma \cdot \bar{t}$ (s)} & \multicolumn{1}{c}{$\overline{\text{Gap}}$ (\%)} & \#Wins & \#Ties & \#Losses & \emph{p}-value \\  \hline
\multirow{10}{*}{MDVRP} & \multirow{7}{*}{\emph{C97}} & \multirow{7}{*}{33} & BKS & 2423.02 & - & - & 0.00 & 7 & 23 & 3 & 0.047 \\
 &  &  & CGL & 2455.56 & 1112.27 & - & 1.16 & 25 & 8 & 0 & 1.23E-05 \\
 &  &  & PR & 2428.30 & 236.61 & 63.88 & 0.21 & 15 & 18 & 0 & 9.82E-04 \\
 &  &  & VCGLR & 2425.22 & 254.30 & 68.66 & 0.10 & 8 & 25 & 0 & 0.012 \\
 &  &  & ELTG & 2430.12 & 138.82 & 43.03 & 0.29 & 11 & 19 & 3 & 3.51E-03 \\
 &  &  & SCA & 2430.86 & 217.37 & 293.45 & 0.20 & 15 & 17 & 1 & 6.43E-04 \\
 &  &  & MDFIHA & 2421.33 & 229.61 & 229.61 & -0.09 & - & - & - & - \\ \cline{2-12} 
 & \multirow{3}{*}{\emph{C97-T}} & \multirow{3}{*}{10} & Optima* & 1879.31 & - & - & 0.00 & 0 & 10 & 0 & - \\
 &  &  & SCA & 1881.16 & 380.64 & 513.86 & 0.08 & 3 & 7 & 0 & 0.109 \\
 &  &  & MDFIHA & 1879.31 & 382.19 & 382.19 & 0.00 & - & - & - & - \\ \hline
\multirow{9}{*}{MDVRPTW} & \multirow{4}{*}{\emph{C01}} & \multirow{4}{*}{20} & BKS & 2224.42 & - & - & 0.00 & 4 & 15 & 1 & 0.138 \\ 
 &  &  & VCGP & 2224.43 & 389.58 & 280.50 & 0.00 & 5 & 14 & 1 & 0.075 \\
 &  &  & SCA & 2225.91 & 1022.86 & 1380.86 & 0.05 & 9 & 11 & 0 & 0.012 \\
 &  &  & MDFIHA & 2223.61 & 732.71 & 732.71 & -0.03 & - & - & - & - \\ \cline{2-12}  	
 & \multirow{2}{*}{\emph{C01-R}} & \multirow{2}{*}{10} & VCGP & 2021.60 & 451.98 & 325.43 & 0.00 & 9 & 1 & 0 & 0.008 \\
 &  &  & MDFIHA & 1988.86 & 554.29 & 554.29 & -2.14 & - & - & - & - \\ \cline{2-12} 
 & \multirow{3}{*}{\emph{V13}} & \multirow{3}{*}{28} & VCGP & 8779.76 & 5338.86 & 3843.98 & 0.00 & 19 & 0 & 9 & 0.048 \\
 &  &  & MDFIHA & 8765.20 & 6352.19 & 6352.19 & -0.22 & - & - & - & - \\
 &  &  & MDFIHA-ETGA \dag & 8722.38 & 6490.46 & 6490.46 & -0.60 & 26 & 0 & 2 & 6.56E-07 \\ \hline
\multirow{5}{*}{MDOVRP} & \multirow{5}{*}{\emph{L14}} & \multirow{5}{*}{24} & BKS & 1398.69 & - & - & 0.00 & 6 & 17 & 1 & 0.028 \\
 &  &  & JB & 1402.74 & 64.57 & 58.11 & 0.22 & 15 & 9 & 0 & 6.55E-04 \\
 &  &  & RM & 1411.14 & 2551.24 & 2576.75 & 0.51 & 7 & 17 & 0 & 0.018 \\
 &  &  & SCA & 1399.38 & 141.44 & 190.94 & 0.03 & 6 & 17 & 1 & 0.028 \\
 &  &  & MDFIHA & 1397.39 & 78.99 & 78.99 & -0.06 & - & - & - & - \\ \hline 		
\end{tabular}%
}
}{\parbox[t]{\textwidth}{\raggedright\footnotesize
*Optimal solutions from \cite{sadykov2021bucket} are used as references for the \emph{C97-T} instances. \\ 
\dag The comparisons and \emph{p}-values for MDFIHA-ETGA refer to its performance relative to VCGP.
}}
\end{table}

On the MDVRP \emph{C97} instances, MDFIHA achieves an average gap of -0.09\%, establishes 7 new best upper bounds, and matches the BKS for 23 of the remaining 26 instances. While some older algorithms show shorter running times, especially after time conversion, MDFIHA consistently delivers better solution quality across most instances, and it outperforms the recent algorithm SCA \cite{sadati2021efficient} in both quality and computation time. On the \emph{C97-T} instances with tighter fleet-size limits, MDFIHA reaches all 10 known optimal solutions with shorter computation times than SCA. 

On the well-solved MDVRPTW \emph{C01} instances, MDFIHA achieves an average gap of -0.03\%, establishes 4 improved upper bounds, and again outperforms the recent SCA. Against the high-performing VCGP algorithm \cite{vidal2013hybrid}, MDFIHA achieves better solution quality on 5 instances and it also finds 9 better solutions out of 10 \emph{C01-R} instances with relaxed fleet-size limits.

On the 28 large-scale and challenging MDVRPTW \emph{V13} instances, despite using a shorter time limit (7200 seconds) compared to VCGP's 5-hour limit (18000 seconds, converted to 12960 seconds), MDFIHA achieves an average gap of -0.22\% and improves 19 best-known solutions. The GPU-enhanced MDFIHA-ETGA delivers even stronger results, with an average gap of -0.60\% and 26 improved upper bounds, highlighting its scalability and efficiency when addressing large-scale instances.

Finally, MDFIHA outperforms all three reference algorithms on the MDOVRP \emph{L14} instances, achieving an average gap of -0.05\%, reporting 6 improved upper bounds within short computation times. These results further demonstrate the effectiveness of the proposed algorithm across different problem variants.

Additionally, most associated \emph{p}-values are below 0.05, indicating that MDFIHA statistically outperforms both the BKS and the reference algorithms. Exceptions on the \emph{C97-T} and \emph{C01} sets are due to small benchmark sizes and the presence of known optimal solutions or well-solved instances, for which no further improvement is possible.

\section{Performance analysis}
\label{sec:analysis}

This section analyzes the impact of DCREX crossover and MDFIS procedure on the performance of the proposed algorithm on the 73 benchmark instances of \emph{C97}, \emph{C97-T}, \emph{C01}, and \emph{C01-R}. To evaluate the impact of GPU acceleration within the ETGA framework, we use the 28 large-scale \emph{V13} instances.

\subsection{Analysis on diversity-controlled route-exchange crossover}
\label{sec:analysis_crossover}

To investigate the impact of the DCREX crossover, we construct several variants based on the original MDFIHA algorithm as the baseline:
\begin{itemize}
    \item MDFIHA-RDP: Remove DCREX and use a single-individual population. Each generation applies a random destroy-and-repair operator followed by local optimization with MDFIS.
    \item MDFIHA-CDP: The same as MDFIHA-RDP, but we use a correlated destroy-and-repair operator based on the correlation metric from Section \ref{sec:pruning}.
    \item MDFIHA-SREX: Replace DCREX with the selective route exchange crossover (SREX) \cite{nagata2010memetic}, which swaps correlated route clusters between parents.
    \item MDFIHA-RARI: Replace DCREX with the route assembly regret insertion crossover (RARI) \cite{liu2021memetic}, which iteratively inherits routes from two parents and then repairs the offspring.
\end{itemize}
\begin{figure}[!htbp]
\centering
\includegraphics[width=0.5\textwidth]{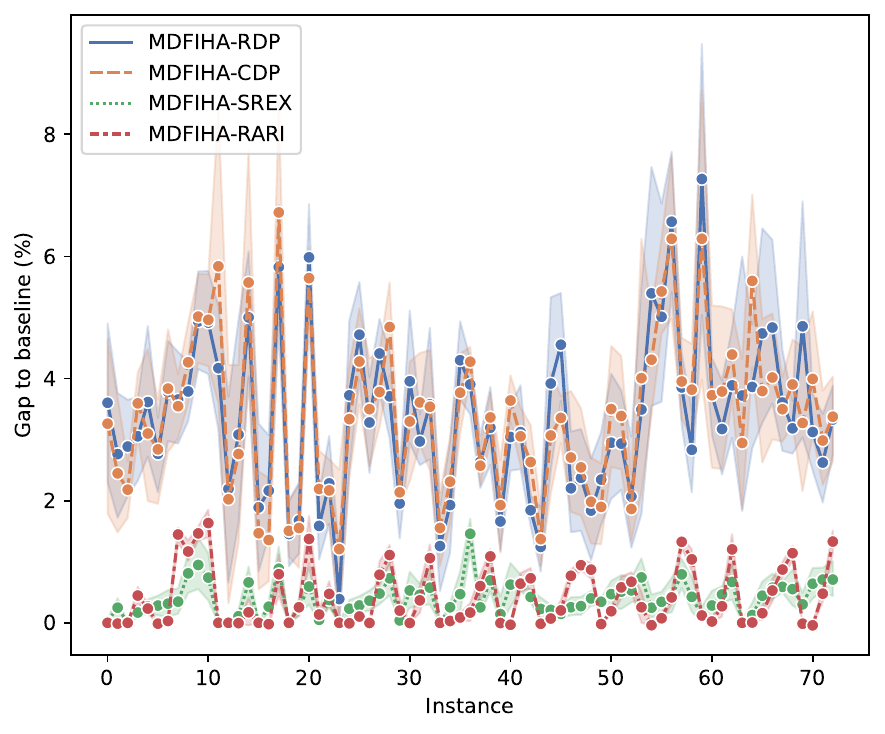}
\caption{Performance comparison of MDFIHA and its variants. The x-axis indicates instances, and the y-axis shows the relative gap to the baseline MDFIHA. Colored lines denote mean gaps with 0.95 confidence intervals.}
\label{fig:ana-x}
\end{figure}

Figure \ref{fig:ana-x} compares MDFIHA with its variants using different destroy-and-repair and crossover strategies. MDFIHA-RDP and MDFIHA-CDP, which rely on single-solution destroy-and-repair methods, perform significantly worse, highlighting the advantage of population-based search. While MDFIHA-SREX and MDFIHA-RARI outperform the destroy-and-repair-based variants, they still lag behind the original MDFIHA, confirming the effectiveness of the DCREX crossover.

\begin{figure}[!htbp]
    \centering
    \begin{subfigure}[t]{0.45\textwidth}
        \centering
        \includegraphics[width=\textwidth]{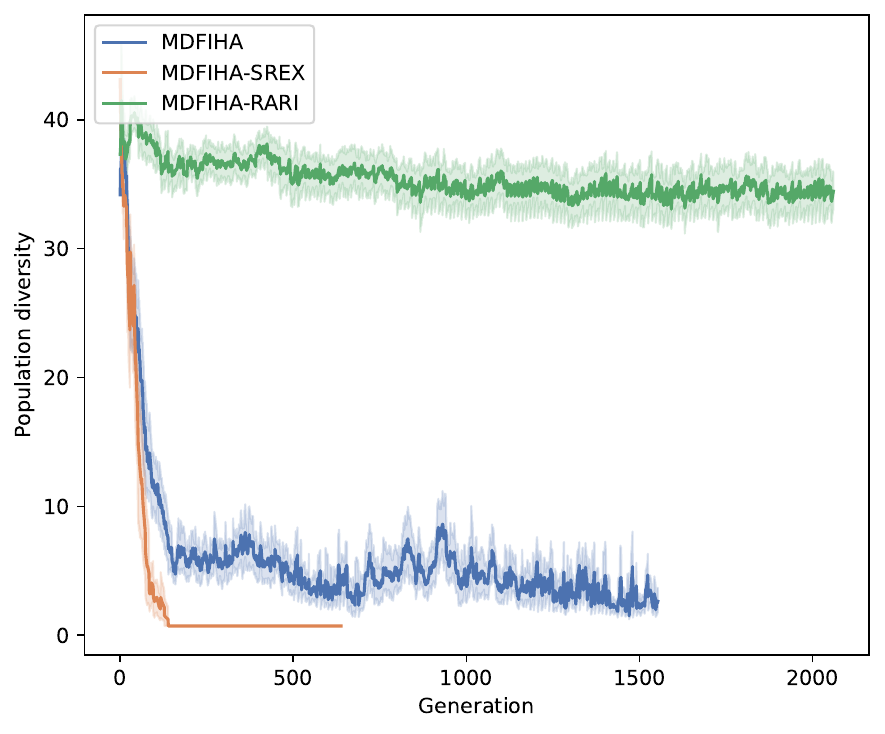}
        \caption{Instance \emph{p11} from \emph{C97}.}
        \label{fig:ana-pop-c97-p11}
    \end{subfigure}
    \hfill
    \begin{subfigure}[t]{0.45\textwidth}
        \centering
        \includegraphics[width=\textwidth]{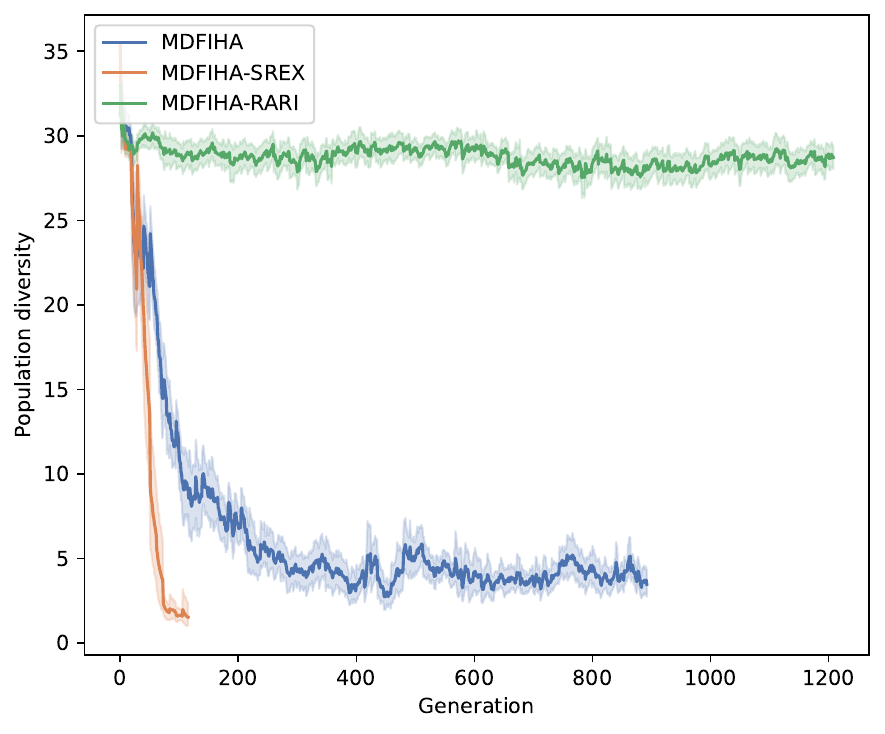}
        \caption{Instance \emph{p21} from \emph{C97}.}
        \label{fig:ana-pop-c97-p21}
    \end{subfigure}
    \hfill
    \begin{subfigure}[t]{0.45\textwidth}
        \centering
        \includegraphics[width=\textwidth]{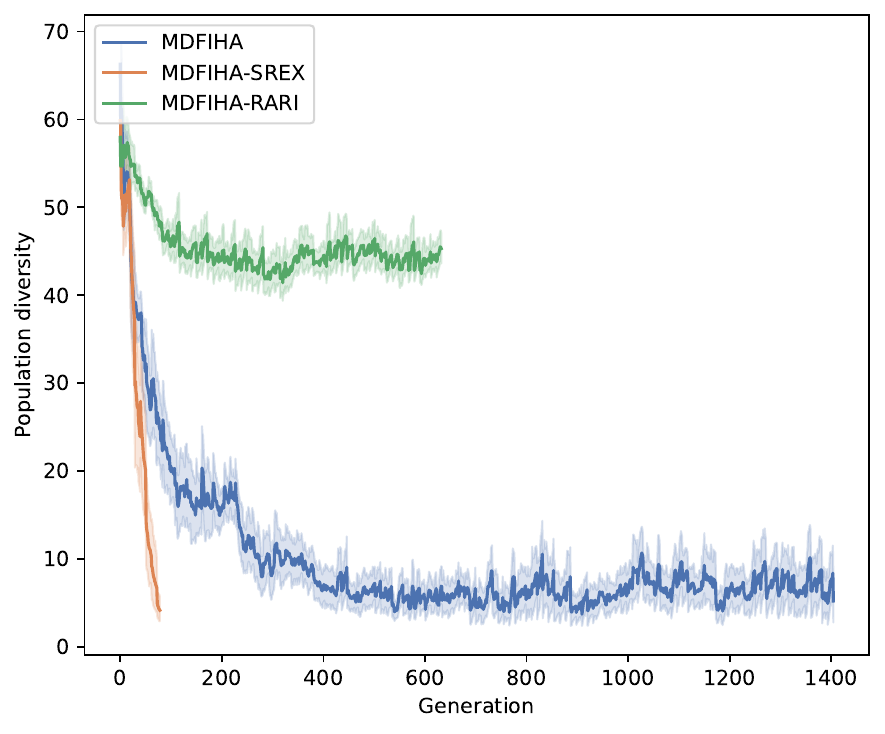}
        \caption{Instance \emph{pr05} from \emph{C01}.}
        \label{fig:ana-pop-c01-pr05}
    \end{subfigure}
    \hfill
    \begin{subfigure}[t]{0.45\textwidth}
        \centering
        \includegraphics[width=\textwidth]{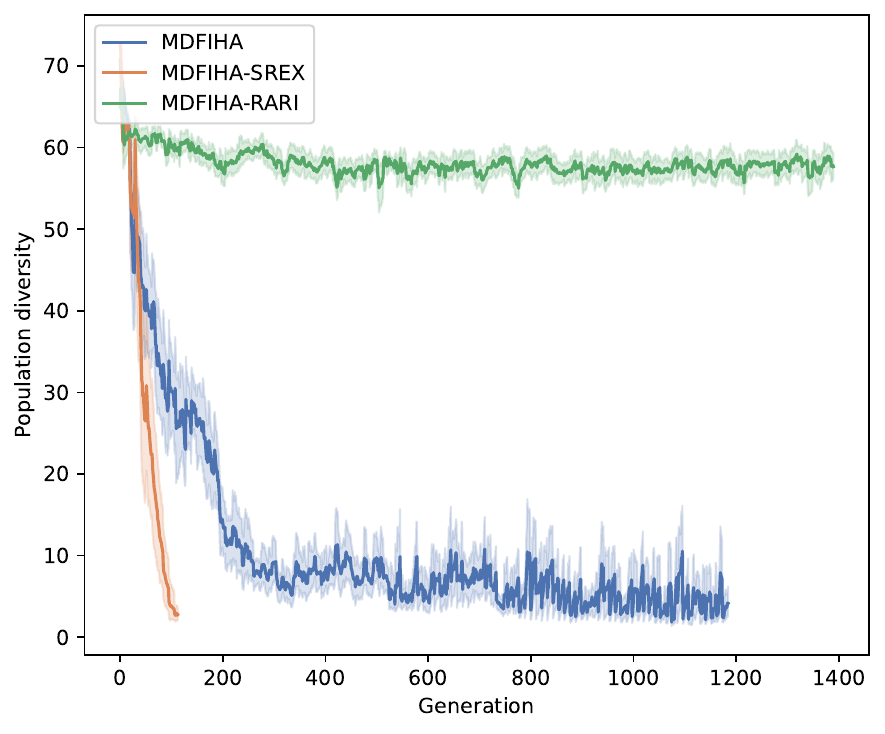}
        \caption{Instance \emph{pr15} from \emph{C01}.}
        \label{fig:ana-pop-c01-pr15}
    \end{subfigure}
    \caption{Evolution of population diversity across generations for MDFIHA and its variants MDFIHA-SREX and MDFIHA-RARI. The diversity is measured by the solution distances within population.}
    \label{fig:ana-pop-diversity}
\end{figure}

To better understand the role of DCREX, we examine its impact on population diversity throughout the search process. Figure \ref{fig:ana-pop-diversity} illustrates the evolution of population diversity for four representative instances. In all cases, MDFIHA-SREX exhibits a rapid decrease in diversity, resulting in premature convergence. Although MDFIHA-RARI maintains a high level of diversity all the time, it fails to converge to high-quality solutions. In contrast, the proposed DCREX crossover enables faster convergence in the early stages while preserving a moderate level of diversity later on. This helps the algorithm to balance exploration and exploitation more effectively, leading to better overall results.

\subsection{Analysis on multi-depot-supported feasible-and-infeasible search}
\label{sec:analysis_search}

To study the impact of the multi-depot-supported feasible-and-infeasible search, we construct the following MDFIHA variants. 

\begin{itemize}
    \item MDFIHA-F: Accept only feasible solutions during the search.
    \item MDFIHA-PC10: Retain feasible and infeasible search with a fixed penalty coefficient of 10.
    \item MDFIHA-wo-DODT: Remove \emph{Depot-Insert} and \emph{Depot-Replace} operators and excludes the depot-related penalty from the evaluation function (Equation (\ref{eq:eval_func})).
\end{itemize}

\begin{figure}[!htbp]
\centering
\includegraphics[width=0.5\textwidth]{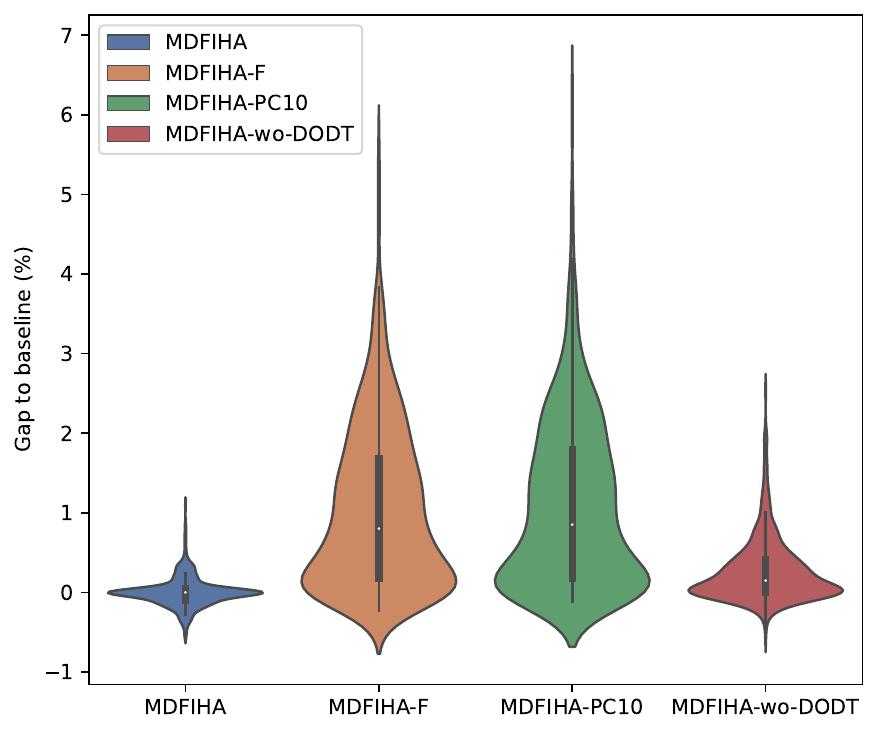}
\caption{Performance comparison of MDFIHA and its variants. The x-axis indicates algorithm variants, and the y-axis shows the relative gap to the baseline MDFIHA.}
\label{fig:ana-search}
\end{figure}

According to the violin plots in Figure \ref{fig:ana-search}, MDFIHA-F and MDFIHA-PC10 perform significantly worse than the baseline MDFIHA, with notably higher median gaps, greater variance and more extreme outliers. This suggests that either excluding infeasible solutions or using fixed penalty coefficients reduces the effectiveness of the search process. Similarly, MDFIHA-wo-DODT also shows degraded performance, which confirms the importance of depot-related components in improving the quality of solutions in multi-depot problems.

\begin{figure}[!htbp]
    \centering
    \begin{subfigure}[t]{0.48\textwidth}
        \centering
        \includegraphics[width=\textwidth]{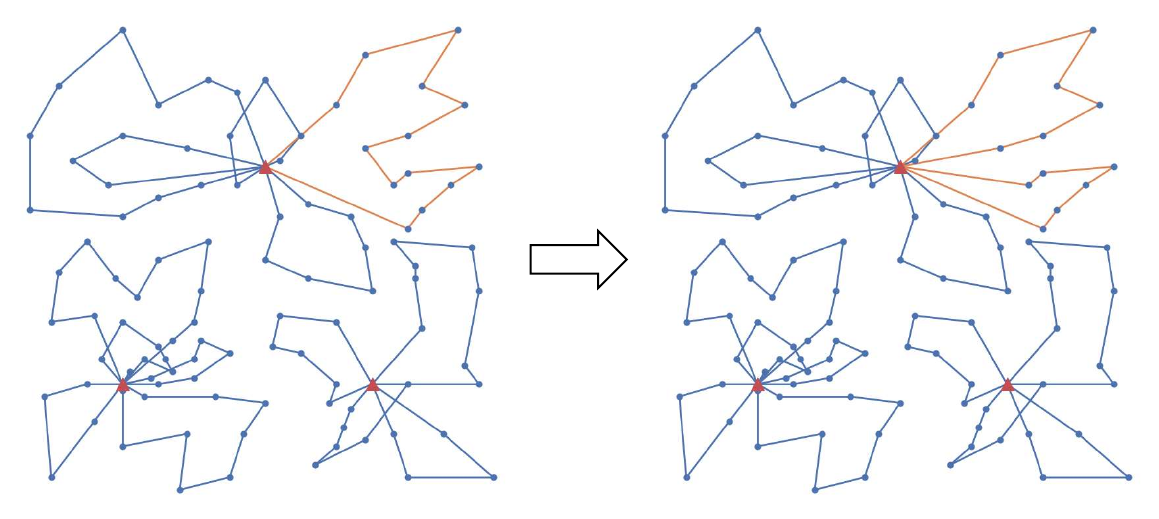}
        \caption{A \emph{Depot-Insert} move on instance \emph{p06} from \emph{C97}.}
        \label{fig:illu-insert}
    \end{subfigure}
    \hfill
    \begin{subfigure}[t]{0.48\textwidth}
        \centering
        \includegraphics[width=\textwidth]{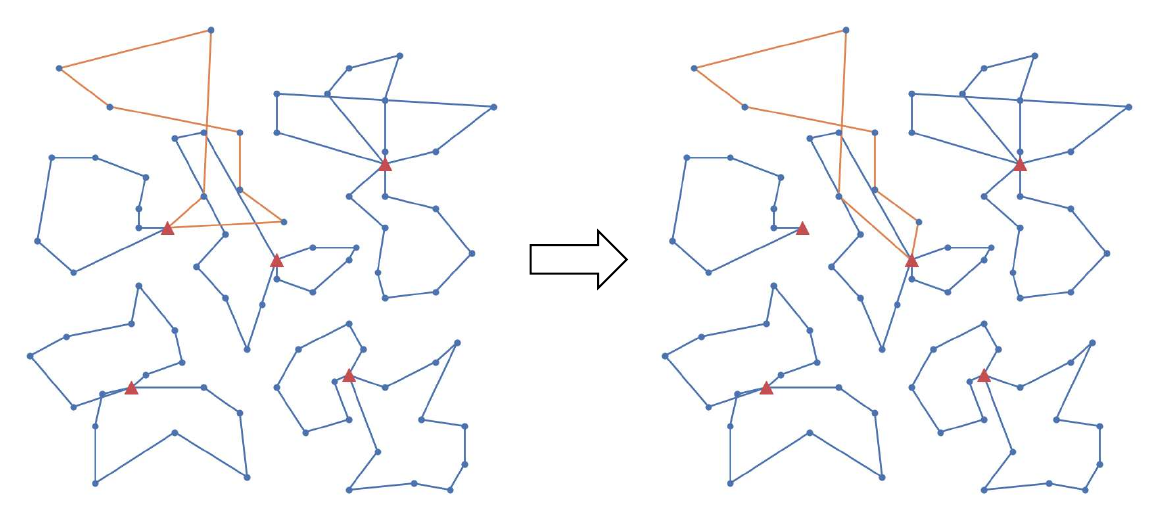}
        \caption{A \emph{Depot-Replace} move on instance \emph{p03} from \emph{C97}.}
        \label{fig:illu-replace}
    \end{subfigure}
    \caption{Illustration of depot-related operators. Red triangles denote depots, blue points represent customers. Blue edges indicate unchanged routes, yellow edges highlight modified routes.}
    \label{fig:illu-depot-opts} 
\end{figure}

Additionally, to clarify the roles of the depot operators, Figure \ref{fig:illu-depot-opts} illustrates the actions of \emph{Depot-Insert} and \emph{Depot-Replace} during the search. \emph{Depot-Insert} is observed to effectively split a route into two, enabling rapid transitions in the search space, while \emph{Depot-Replace} operator moves an entire route to a different depot, which is essentially a depot reallocation. These two operators jointly enhance the algorithm's ability to explore the multi-depot solution space more thoroughly by simultaneously considering both depot assignments and vehicle routings.

\subsection{Analysis of GPU parallelism benefits}
\label{sec:analysis_gpu}

To evaluate the benefits of GPU parallelism, including the computational efficiency and speedup achieved by the edge-implemented tensor-based GPU acceleration, as well as the performance gains from the proposed leader-follower multi-move update strategy, we conduct additional experiments on the 28 large-scale MDVRPTW \emph{V13} instances. 

\begin{figure}[!htbp]
    \centering
    \begin{subfigure}[t]{0.48\textwidth}
        \centering
        \includegraphics[width=\textwidth]{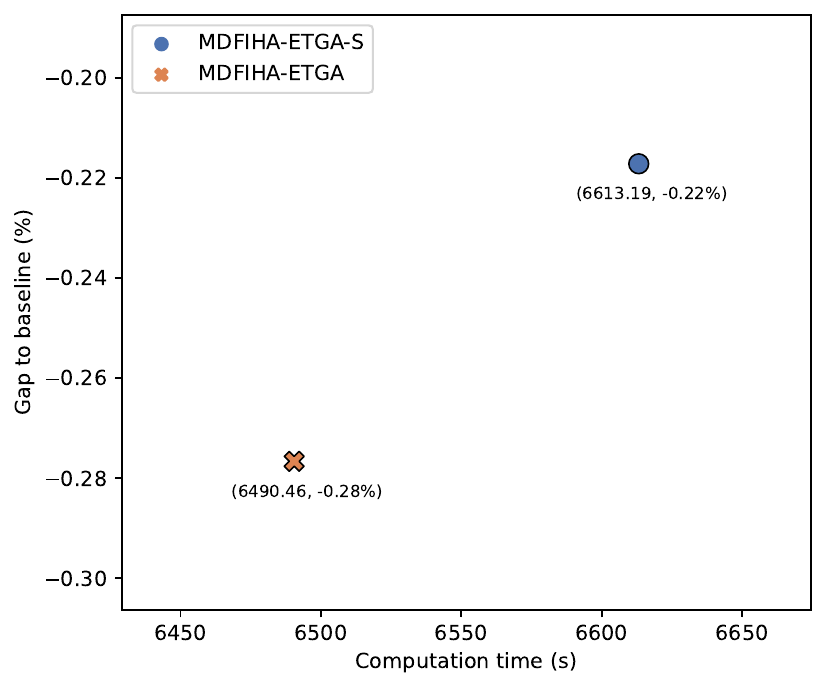}
        \caption{Overall performance comparison. The x-axis reports the average computation time in seconds, and the y-axis shows the relative gap to the baseline.}
        \label{fig:ana-mm-overall}
    \end{subfigure}
    \hfill
    \begin{subfigure}[t]{0.48\textwidth}
        \centering
        \includegraphics[width=\textwidth]{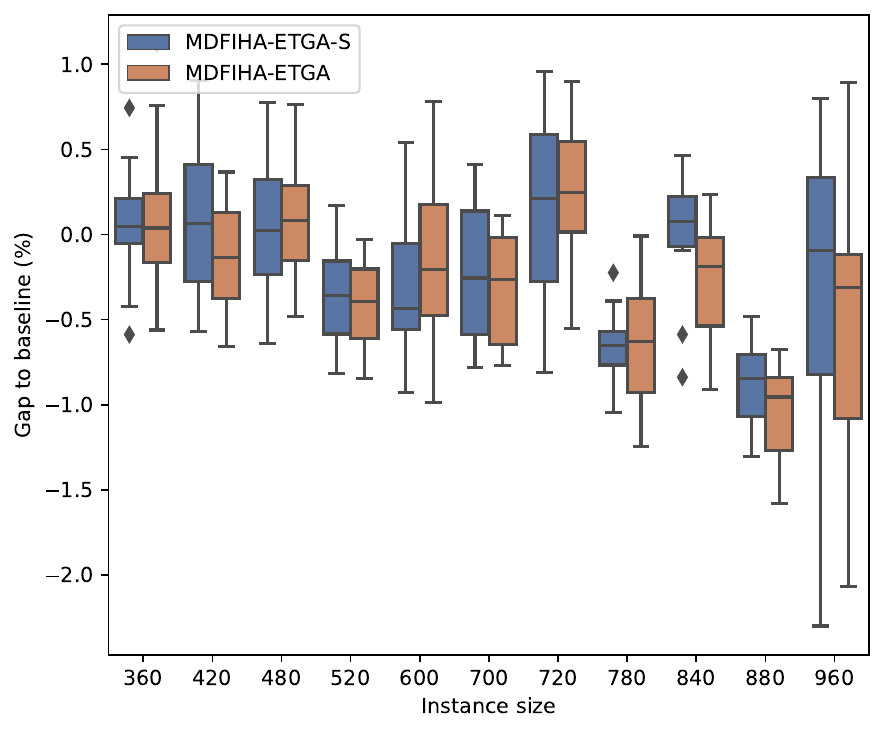}
        \caption{Performance comparison across instance sizes. The x-axis denotes instance size, and the y-axis shows the relative gap to the baseline.}
        \label{fig:ana-mm-inst-size}
    \end{subfigure}
    \caption{Performance comparison between MDFIHA-ETGA and MDFIHA-ETGA-S, with VCGP \cite{vidal2013hybrid} serving as the baseline.}
    \label{fig:ana-multi-move} 
\end{figure}

First, we compare MDFIHA-ETGA and MDFIHA-ETGA-S under the same experimental settings as set in Section \ref{sec:exp_protocol}, where MDFIHA-ETGA-S is a variant that adopts a conventional single-move mechanism without the multi-move strategy of Section \ref{sec:multi_move}. This comparison aims to isolate and assess the contribution of the proposed multi-move update strategy. The corresponding results are reported in Figure \ref{fig:ana-multi-move}.

Figure \ref{fig:ana-mm-overall} shows that MDFIHA-ETGA achieves better overall performance than MDFIHA-ETGA-S, with a lower average gap and shorter average computation time. This observation highlights the effectiveness of the proposed leader-follower multi-move update strategy in enhancing solution quality while preserving computational efficiency. Figure \ref{fig:ana-mm-inst-size} further presents the performance comparison across different instance sizes. While some variability can be observed, MDFIHA-ETGA demonstrates more stable and robust behavior and attains notably better average solution quality, especially for large-scale instances with more than 800 nodes, compared to MDFIHA-ETGA-S. These results suggest that the leader-follower multi-move strategy is particularly well suited for larger and more complex instances, as it enables more effective exploration of the search space and a greater ability to escape local optima than conventional single-move updates.

\begin{figure}[!htbp]
    \centering
    \begin{subfigure}[t]{0.45\textwidth}
        \centering
        \includegraphics[width=\textwidth]{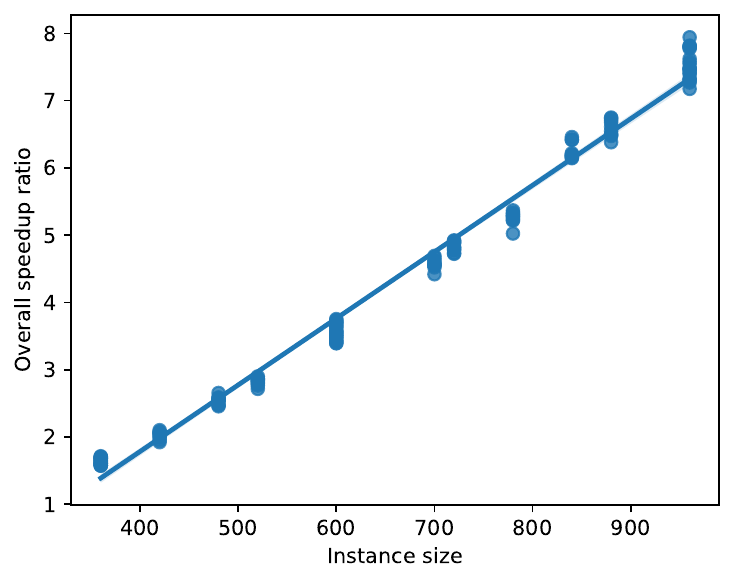}
        \caption{Overall speedup across different instance sizes, with a regression slope $\approx$ 0.0099 and $R^2 = 0.99$.}
        \label{fig:ana-overall-speedup}
    \end{subfigure}
    \hfill
    \begin{subfigure}[t]{0.45\textwidth}
        \centering
        \includegraphics[width=\textwidth]{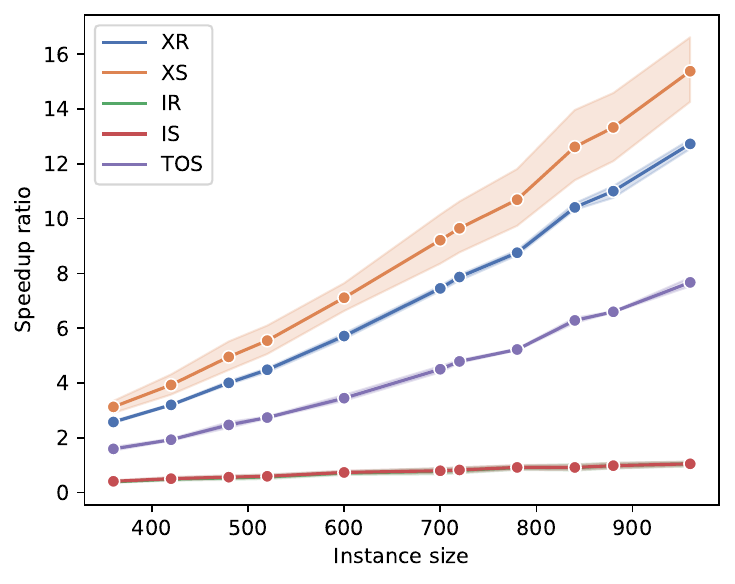}
        \caption{Speedup ratios for operators across different instance size.}
        \label{fig:ana-o-speedup}
    \end{subfigure}
    \caption{ETGA speedup analysis on \emph{V13} instances. The x-axis indicates the instance sizes, y-axis represents the speedup ratio.}
    \label{fig:ana-etga} 
\end{figure}

Next, to independently evaluate the computational efficiency in move evaluation of ETGA, we compare MDFIHA-ETGA-S with the CPU-based MDFIHA, both employing the same single best move strategy. Each algorithm is run five times per instance, with each run limited to 100 generations. The overall speedup ratio is defined as $\gamma_s = \frac{t_{\text{MDFIHA}}}{t_{\text{MDFIHA-ETGA-S}}}$, where $t_{\text{MDFIHA}}$ and $t_{\text{MDFIHA-ETGA-S}}$ denote the average computation times for move evaluation in MDFIHA and MDFIHA-ETGA-S, respectively. Speedups for individual operators are computed similarly, based on their average computation times. The experimental results are shown in Figure \ref{fig:ana-etga}.

Figure \ref{fig:ana-overall-speedup} shows a strong linear relationship (coefficient of determination $R^2=0.99$) between instance size and speedup, with slope $\approx$ 0.0099. This indicates excellent scalability of the ETGA, which means that larger instances benefit from greater acceleration. Specifically, the speedup grows from about 1.6 for 360-node instances to nearly 7.5 for 960-node instances, demonstrating significant improvements in computational efficiency, even under neighborhood reduction and pruning settings.

Figure \ref{fig:ana-o-speedup} illustrates the speedup ratios of different local search operators across various instance sizes in the \emph{V13} benchmark. Recall that the operators in ETGA are categorized as intra-operators (\emph{Intra-Relocate} (IR), \emph{Intra-Swap} (IS), and  \emph{2-opt} (TO)) and inter-operators (\emph{Inter-Relocate} (XR), \emph{Inter-Swap} (XS), and \emph{2-opt*} (TOS)) depending on whether they act on one or two routes. It can be observed that inter-operators (XR, XS, TOS) achieve substantial speedups because the ETGA framework efficiently parallelizes their computationally intensive evaluations on the GPU. In contrast, intra-route operators (IR, IS) on a single route show more modest speedups. This is because these operators have lower time complexity and benefit less from GPU parallelism; the overhead of the GPU operations becomes more significant relative to the total computation.

\section{Conclusion}
\label{sec:conclusion}

We propose a novel hybrid algorithm MDFIHA for solving a class of multi-depot vehicle routing problems. MDFIHA incorporates several key innovations: a multi-depot-supported feasible-and-infeasible search guided by a multi-penalty evaluation function and enhanced with two depot-related operators to effectively address the complex multi-depot constraints, and a learning-driven diversity-controlled route-exchange crossover to generate promising offspring solutions. To further improve computational efficiency on large-scale instances, we develop the enhanced MDFIHA-ETGA algorithm that integrates the edge-implemented tensor-based GPU acceleration to significantly accelerate neighborhood evaluations, and incorporates a leader-follower multi-move update strategy to fully exploit GPU parallelism and enhance solution quality.

Extensive experiments on four benchmark sets, covering three different MDVRP problems and totaling 125 instances, demonstrate the effectiveness of the proposed algorithm. MDFIHA achieves competitive or superior performance compared to state-of-the-art algorithms, establishing 45 record-breaking upper bounds. Additionally, the tensor-based GPU acceleration framework with multi-move update strategy in MDFIHA-ETGA significantly enhances the search process, achieving substantial speedups while further improving solution quality on large-scale instances.

The proposed algorithm has strong potential for extension to other vehicle routing problems. Future work will investigate these extensions and further leverage the tensor-based GPU acceleration framework and GPU parallelism to enhance scalability and computational efficiency.

\section*{Acknowledgments}
\label{sec:acknowledgments}

This work was granted access to the HPC resources of IDRIS under the allocation 2025-AD010617072 made by GENCI. We also acknowledge the HPC resources provided by the Centre de Calcul Intensif des Pays de la Loire (CCIPL), France.

\bibliographystyle{elsarticle-num}
\bibliography{references}

\newpage         
\appendix             

\renewcommand{\thesection}{A\arabic{section}}
\setcounter{section}{0}
\renewcommand{\thetable}{A\arabic{table}}
\setcounter{table}{0}
\renewcommand{\thefigure}{A\arabic{figure}}
\setcounter{figure}{0}

\section*{Supplementary materials} 

The supplementary materials include a detailed definition of the multi-penalty evaluation function, an illustration of the diversity-controlled route-exchange crossover (DCREX), and detailed computational results on benchmark MDVRP instances.

\section{Multi-penalty evaluation function}
\label{app:eval_func}

The multi-penalty evaluation function used by the multi-depot-supported feasible-and-infeasible search (MDFIS) procedure is shown in Equation (\ref{eq:eval_func_sm}). This function aims to minimize the total travel distance while integrating four normalized penalty terms corresponding to violations of time windows, vehicle capacity, maximum route duration, and depot-related constraints. 

\begin{align} 
    \label{eq:eval_func_sm}
    \mathrm{Minimize} \quad \mathcal{F}(S) &= D(S) + \lambda_1 \cdot V_{1}(S) + \lambda_2 \cdot V_{2}(S) + \lambda_3 \cdot V_{3}(S) + \lambda_4 \cdot V_{4}(S)  \\
    \mathrm{where} \quad
    D(S) &= \sum_{i=1}^{M} \sum_{j=0}^{L_i} c_{n_{i,j}n_{i,j+1}} \nonumber \\
    V_{1}(S) &= \Gamma \cdot \sum_{i=1}^{M} \sum_{j=1}^{L_i} \max\{a_{n_{i,j}} - l_{n_{i,j}}, 0\} \nonumber \\
    V_{2}(S) &= \Theta \cdot \frac{ \sum_{i=1}^{M} \max\{\sum_{j=1}^{L_i} q_{n_{i,j}} - \mathcal{Q}, 0\}}{\mathcal{Q}} \nonumber \\
    V_{3}(S) &= \Theta \cdot \frac{ \sum_{i=1}^{M} \max\{\sum_{j=0}^{L_i} (w_{n_{i,j}} + s_{n_{i,j}} + t_{n_{i,j}n_{i,j+1}}) - \mathcal{D}, 0\}}{\mathcal{D}} \nonumber \\
    V_{4}(S) &= \Theta \cdot (\sum_{i=1}^{M} 1_{\{n_{i,0} \ne n_{i,L_{i+1}}\}} + \sum_{n_d \in \mathcal{V}_D} \max\{\sum_{i=1}^{M} 1_{\{n_{i,0} = n_d\}} - N_V, 0\}) \nonumber \\
    S &= \{R_1, \dots, R_M\} \nonumber \\
    R_i &= \{n_{i,0}, n_{i,1}, \dots, n_{i,L_i}, n_{i,L_i+1}\}, \quad i =1,\dots,M  \nonumber \\
    \Gamma &= \frac{\sum_{e_{uv}\in \mathcal{E}} c_{uv}}{\sum_{e_{uv}\in \mathcal{E}} t_{uv}} \nonumber \\
    \Theta &= 2 \cdot \max c_{uv} - \min c_{uv}, \qquad e_{uv}\in \mathcal{E} \nonumber 
\end{align}

In this formulation, $D(S)$ denotes the total distance traveled by all vehicles, which is the primary objective of the MDVRPs. The terms $V_{1}(S)$, $V_{2}(S)$, $V_{3}(S)$, and $V_{4}(S)$ represent the normalized violation values for the time windows, vehicle capacity, maximum tour duration, and depot related constraints, respectively. The factor $\Gamma$ represents the average distance-time ratio over all edges and is used to scale the time-window violation term into distance units. A scaling factor $\Theta$ is introduced to normalize the penalty terms, calculated based on the maximum and minimum edge distances. To handle violations of time window constraints, we adopt the time-warp technique \cite{vidal2013hybrid}, which allows for a smooth penalty application when time constraints are violated. The depot-related penalty term $V_{4}(S)$ consists of two components. The first is the route closure violation, which penalizes routes that do not start and end at the same depot. The second is the depot capacity violation, which penalizes solutions where the number of vehicles assigned to a depot exceeds its available vehicle limit.

\section{Illustration of the DCREX crossover}
\label{app:crossover}
\begin{figure}[htbp]
    \centering
    \begin{subfigure}[t]{0.75\textwidth}
        \centering
        \includegraphics[width=\textwidth]{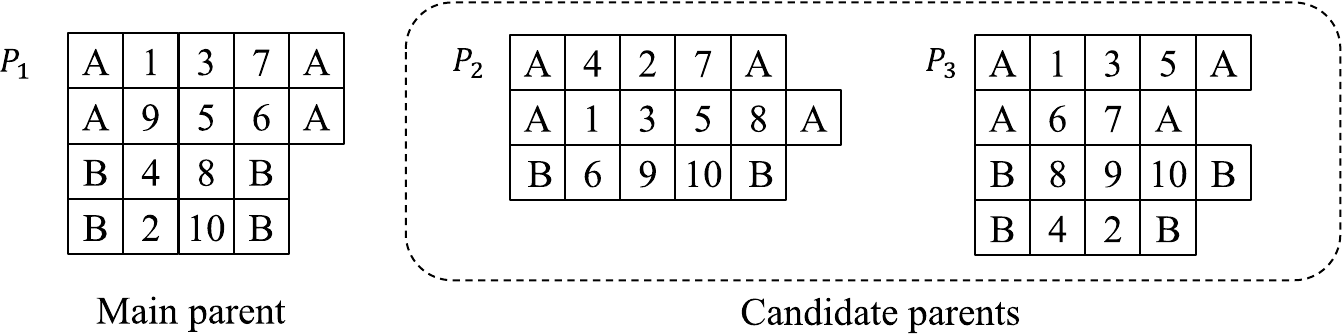}
        \caption{Main parent $P_1$ and candidate parents $P_2$ and $P_3$.}
        \label{fig:dcrex1}
    \end{subfigure}
    \hfill
    \begin{subfigure}[t]{0.75\textwidth}
        \centering
        \includegraphics[trim=5 5 5 5, clip,
            width=\textwidth]{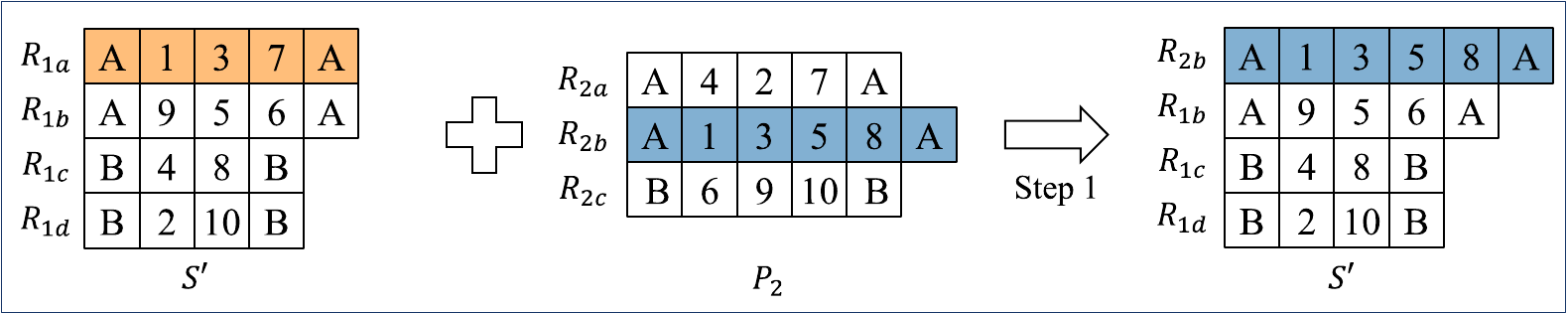}
        \caption{Selected route pair $\{R_{1a}, R_{2b}\}$ with score computed as $-(E^I - N^R - N^M - 10 \cdot N^C) = -(3 - 2 - 1 - 0) = 0$, and diversity increase $\Delta \sigma = E^I + N^R + N^M = 6$.}
        \label{fig:dcrex2}
    \end{subfigure}
    \hfill
    \begin{subfigure}[t]{0.75\textwidth}
        \centering
        \includegraphics[trim=5 5 5 5, clip,
            width=\textwidth]{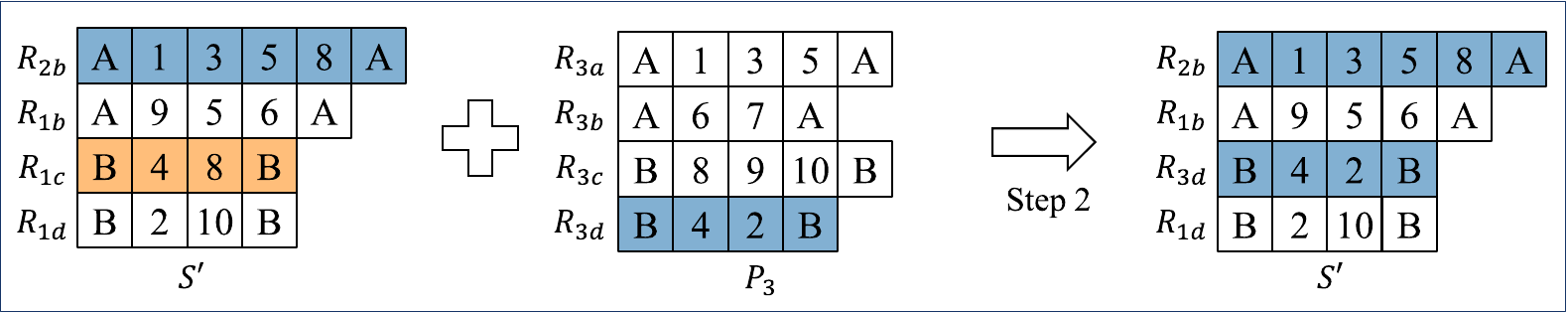}
        \caption{Selected route pair $\{R_{1c}, R_{3d}\}$ with score computed as $-(E^I - N^R - N^M - 10 \cdot N^C) = -(1 - 0 - 0 - 0) = -1$, and diversity increase $\Delta \sigma = E^I + N^R + N^M = 1$.}
        \label{fig:dcrex3}
    \end{subfigure}
    \hfill
    \begin{subfigure}[t]{0.75\textwidth}
        \centering
        \includegraphics[trim=5 5 5 5, clip, 
            width=\textwidth]{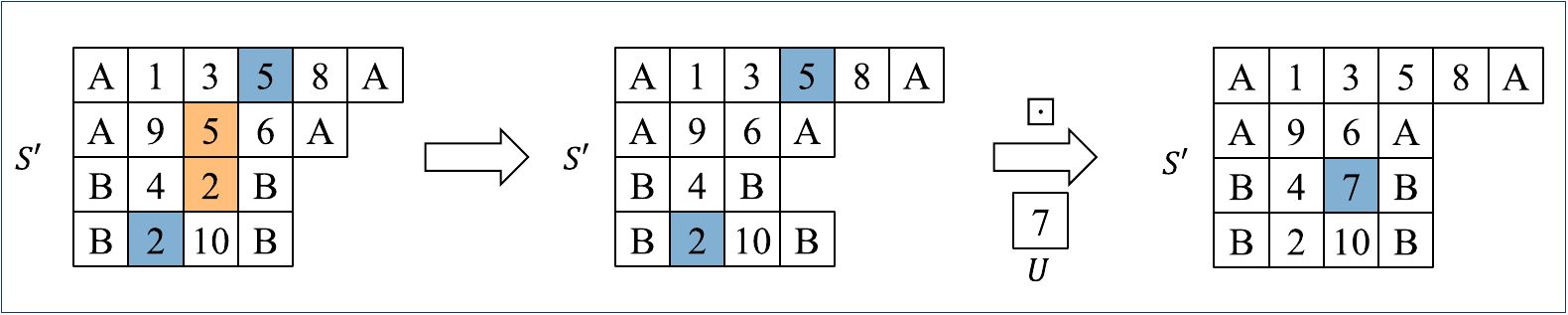}
        \caption{Redundant nodes removal and unrouted nodes insertion.}
        \label{fig:dcrex4}
    \end{subfigure}
    \caption{Illustration of the diversity-controlled route-exchange crossover. ``A" and ``B" are depot nodes. Blue indicates the routes added or nodes retained, while yellow represents the routes or nodes that are removed.}
    \label{fig:dcrex} 
\end{figure}

Figure \ref{fig:dcrex} shows a step-by-step example of the diversity-controlled route-exchange crossover (DCREX) proposed in the paper. This illustration uses three parent solutions, where $P_1$ serves as the main parent $P_m$, while $P_2$ and $P_3$ serve as candidate parents (Figure \ref{fig:dcrex1}). The diversity degree is set to $\sigma_{10} = 0.5$, with corresponding diversity value bounds $\sigma_{\min} = 7.0$ and $\sigma_{\max} = 7.7$. The offspring $S'$ first exchanges a route with $P_2$ (Figure \ref{fig:dcrex2}), yielding a diversity increase of $\Delta \sigma = 6$. Since further exchanges with $P_2$ would exceed $\sigma_{\max}$, $P_3$ is then considered (Figure \ref{fig:dcrex3}), resulting in a second exchange with $\Delta \sigma = 1$, reaching the diversity target $\sigma_{\min} = 7.0$. Finally, redundant nodes 2 and 5 are removed, and the unrouted node 7 is inserted using the selected insertion operator $\boxdot$ (Figure \ref{fig:dcrex4}), producing the final offspring $S'$.

\section{Detailed computational results}
\label{app:computational_results}

We evaluate the proposed algorithms on the classic Multi-Depot Vehicle Routing Problem (MDVRP) and two MDVRP variants: the Multi-Depot Vehicle Routing Problem with Time Windows (MDVRPTW) which introduces time window constraints, and the Multi-Depot Open Vehicle Routing Problem (MDOVRP), in which vehicles are not required to return to their depots after serving customers.

\begin{table}[htbp]
\caption{Benchmark sets.}
\label{tab:benchmark}
\centering
{
\resizebox{0.8\textwidth}{!}{%
\begin{tabular}{llllll}
\hline
Benchmark & Problem & Literature & \#Instances & \#Depots & \#Customers \\ \hline
\emph{C97} & MDVRP & \cite{cordeau1997tabu} & 33 & $[2, 9]$ & $[48, 360]$ \\
\emph{C97-T} & MDVRP & \cite{cordeau1997tabu} & 10 & $[4, 6]$ & $[48, 288]$ \\
\emph{C01} & MDVRPTW & \cite{cordeau2001unified} & 20 & $[4, 6]$ & $[48, 288]$ \\
\emph{C01-R} & MDVRPTW & \cite{cordeau2001unified} & 10 & $[4, 6]$ & $[48, 288]$ \\
\emph{V13} & MDVRPTW & \cite{vidal2013hybrid} & 28 & $[4, 12]$ & $[360, 960]$ \\
\emph{L14} & MDOVRP & \cite{liu2014hybrid} & 24 & $[4, 6]$ & $[48, 288]$
\\ \hline
\end{tabular}%
}
}
\end{table}

The benchmark sets used in our computational experiments are summarized in Table \ref{tab:benchmark}. For each benchmark set, we compare the performance of our algorithms against the best-known solutions (BKS) from the literature and the state-of-the-art algorithms that have contributed to achieving these BKS. Table \ref{tab:ref_algo_sm} recalls the reference algorithms for each benchmark set. Note that the CPU-based MDFIHA algorithm is tested on all benchmark instances, while the GPU-enhanced MDFIHA-ETGA algorithm is tested only on the 28 large-scale \emph{V13} MDVRPTW instances.

\begin{table}[htbp]
\caption{Reference algorithms for benchmark sets.}
\label{tab:ref_algo_sm}
\centering
{
\resizebox{\textwidth}{!}{%
\begin{tabular}{llll}
\hline
Reference algorithm & Approach & Problems & Benchmarks \\
\hline
CGL \cite{cordeau1997tabu} & Tabu Search & MDVRP & \emph{C97} \\
PR \cite{pisinger2007general} & Adaptive Large Neighborhood Search & MDVRP & \emph{C97} \\
VCGLR \cite{vidal2012hybrid} & Hybrid Genetic Search & MDVRP & \emph{C97} \\
ELTG \cite{escobar2014hybrid} & Hybrid Granular Tabu Search & MDVRP & \emph{C97} \\
VCGP \cite{vidal2013hybrid} & Hybrid Genetic Search & MDVRPTW & \emph{C01}, \emph{C01-R}, \emph{V13} \\
JB \cite{brandao2020memory} & Iterated Local Search & MDOVRP & \emph{L14} \\
RM \cite{lalla2021mathematical} & Exact method & MDOVRP & \emph{L14} \\
SCA \cite{sadati2021efficient} & Variable Tabu Neighborhood Search & MDVRP, MDVRPTW, MDOVRP & \emph{C97}, \emph{C97-T}, \emph{C01}, \emph{L14} \\
MDFIHA (Ours) & Memetic Algorithm & MDVRP, MDVRPTW, MDOVRP & \emph{C97}, \emph{C97-T}, \emph{C01}, \emph{C01-R}, \emph{V13}, \emph{L14} \\
MDFIHA-ETGA (Ours) & Memetic Algorithm with GPU acceleration & MDVRPTW & \emph{V13}
\\ \hline
\end{tabular}%
}
}
\end{table}

Our MDFIHA algorithm is run on a computer with an AMD EPYC 7282 2.8 GHz CPU. For the GPU-accelerated algorithm MDFIHA-ETGA, an NVIDIA V100 GPU with 32 GB memory is used. To account for hardware differences, Table \ref{tab:Timing} reports CPU benchmark scores from PassMark (\url{https://www.passmark.com/}) and conversion ratios ($\gamma$) to normalize reported computation times of reference algorithms with our AMD EPYC 7282 as the baseline. The CPU information for the Sun Sparcstation 10 used to run CGL \cite{cordeau1997tabu} is unavailable (indicated by -).

\begin{table}[!htbp]
\caption{Processor information and conversion ratio.}
\label{tab:Timing}
\centering
{
\resizebox{\textwidth}{!}{%
\begin{tabular}{lllllr}
\hline
Reference algorithm & Approach & Processor & Base frequency & \multicolumn{1}{l}{CPU mark} & \multicolumn{1}{l}{$\gamma$} \\ \hline
CGL \cite{cordeau1997tabu} & Tabu Search & Sun Sparcstation 10 & - & - & \multicolumn{1}{l}{-} \\
PR \cite{pisinger2007general} & Adaptive Large Neighborhood Search & Pentium 4 & 3.0 GHz & 528 & 0.27 \\
VCGLR \cite{vidal2012hybrid} & Hybrid Genetic Search & Pentium 4 & 3.0 GHz & 528 & 0.27 \\
ELTG \cite{escobar2014hybrid} & Hybrid Granular Tabu Search & Intel Core Duo & 2.0 GHz & 608 & 0.31 \\
VCGP \cite{vidal2013hybrid} & Hybrid Genetic Search & Intel Xeon & 2.93 GHz & 1404 & 0.72 \\
JB \cite{brandao2020memory} & Iterated Local Search & Intel Core i7-3820 & 3.60 GHz & 1744 & 0.90 \\
RM \cite{lalla2021mathematical} & Exact method & Intel Core i7 & 3.70 GHz & 1960 & 1.01 \\
SCA \cite{sadati2021efficient} & Variable Tabu Neighborhood Search & Intel Core i7-8700 & 3.20 GHz & 2631 & 1.35 \\
MDFIHA (Ours) & Memetic Algorithm & AMD EPYC 7282 & 2.80 GHz & 1945 & 1.00 \\
MDFIHA-ETGA (Ours) & Memetic Algorithm with GPU acceleration & AMD EPYC 7282, NVIDIA V100 &  &  & 1.00
\\ \hline
\end{tabular}%
}
}{}
\end{table}

Tables \ref{tab:exp-md-c97}-\ref{tab:exp-mdo-l14} show the detailed comparison results of the proposed algorithm along with the BKS and the results of the reference algorithms on the benchmark instances. The \emph{Instance} column lists the instance names, and $N_C$, $N_D$, and $N_M$ denote the number of customers, the number of depots, and the number of available vehicles per depot, respectively. The \emph{BKS} column reports the best-known solutions published in the literature, while the `*' symbol or the \emph{Optima} column indicates instances with known optimal values. For each algorithm, we report the best solution found ($f_{best}$), the converted computation time in seconds ($\gamma \cdot t$) using the conversion ratio $\gamma$, and the percentage gap relative to the BKS (Gap(\%)), computed as $\text{Gap}(\%) = \frac{f - \text{BKS}}{\text{BKS}} \times 100$. A negative gap indicates that the algorithm has found a better solution than the BKS, i.e., an improved upper bound. Note that we report the original computation time ($t$) for CGL \cite{cordeau1997tabu}, as its conversion ratio could not be determined due to unavailable CPU benchmark score. For our proposed algorithms, we report the best solution found ($f_{best}$), the average solution ($f_{avg}$) and average computation time ($\gamma \cdot t$) over 10 independent runs. The best solutions are highlighted in bold. The improved upper bounds (a total of 70) are underlined.

\begin{table}[]
\caption{Comparative results on the 33 \emph{C97} MDVRP benchmark instances between BKS, CGL, PR, VCGLR, ELTG, SCA and MDFIHA.}
\label{tab:exp-md-c97}
\resizebox{\textwidth}{!}{%
\begin{tabular}{lrrrrrrcrrcrrcrrcrrcrrrr}
\hline
\multirow{2}{*}{Instance} & \multicolumn{1}{c}{\multirow{2}{*}{$N_C$}} & \multicolumn{1}{c}{\multirow{2}{*}{$N_D$}} & \multicolumn{1}{c}{\multirow{2}{*}{$N_V$}} & \multicolumn{1}{c}{\multirow{2}{*}{BKS}} & \multicolumn{3}{c}{CGL} & \multicolumn{3}{c}{PR} & \multicolumn{3}{c}{VCGLR} & \multicolumn{3}{c}{ELTG} & \multicolumn{3}{c}{SCA} & \multicolumn{4}{c}{MDFIHA (Ours)} \\
\cmidrule(l){6-8} \cmidrule(l){9-11} \cmidrule(l){12-14} \cmidrule(l){15-17} \cmidrule(l){18-20} \cmidrule(l){21-24}
 & \multicolumn{1}{c}{} & \multicolumn{1}{c}{} & \multicolumn{1}{c}{} & \multicolumn{1}{c}{} & \multicolumn{1}{c}{$f_{Best}$} & \multicolumn{1}{c}{$t$} & \multicolumn{1}{c}{Gap (\%)} & \multicolumn{1}{c}{$f_{Best}$} & \multicolumn{1}{c}{$\gamma \cdot t$} & \multicolumn{1}{c}{Gap (\%)} & \multicolumn{1}{c}{$f_{Best}$} & \multicolumn{1}{c}{$\gamma \cdot t$} & \multicolumn{1}{c}{Gap (\%)} & \multicolumn{1}{c}{$f_{Best}$} & \multicolumn{1}{c}{$\gamma \cdot t$} & \multicolumn{1}{c}{Gap (\%)} & \multicolumn{1}{c}{$f_{Best}$} & \multicolumn{1}{c}{$\gamma \cdot t$} & \multicolumn{1}{c}{Gap (\%)} & \multicolumn{1}{c}{$f_{Best}$} & \multicolumn{1}{c}{$f_{Avg}$} & \multicolumn{1}{c}{$\gamma \cdot t$} & \multicolumn{1}{c}{Gap (\%)} \\ \hline
p01 & 50 & 4 & 4 & 576.87* & \textbf{576.87} & 194.00 & \textbf{0.00} & \textbf{576.87} & 7.83 & \textbf{0.00} & \textbf{576.87} & 3.78 & \textbf{0.00} & \textbf{576.87} & 2.17 & \textbf{0.00} & \textbf{576.87} & 33.18 & \textbf{0.00} & \textbf{576.87} & 576.87 & 7.97 & \textbf{0.00} \\
p02 & 50 & 4 & 2 & 473.53* & 473.87 & 208.00 & 0.07 & \textbf{473.53} & 7.56 & \textbf{0.00} & \textbf{473.53} & 3.51 & \textbf{0.00} & \textbf{473.53} & 1.86 & \textbf{0.00} & \textbf{473.53} & 21.03 & \textbf{0.00} & \textbf{473.53} & 473.53 & 18.43 & \textbf{0.00} \\
p03 & 75 & 5 & 3 & 641.19* & 645.15 & 340.00 & 0.62 & \textbf{641.19} & 17.28 & \textbf{0.00} & \textbf{641.19} & 7.02 & \textbf{0.00} & \textbf{641.19} & 8.99 & \textbf{0.00} & \textbf{641.19} & 78.69 & \textbf{0.00} & \textbf{641.19} & 641.60 & 14.01 & \textbf{0.00} \\
p04 & 100 & 2 & 8 & 1001.04* & 1006.66 & 467.00 & 0.56 & \textbf{1001.04} & 23.76 & \textbf{0.00} & \textbf{1001.04} & 31.32 & \textbf{0.00} & \textbf{1001.04} & 27.90 & \textbf{0.00} & \textbf{1001.04} & 85.59 & \textbf{0.00} & \textbf{1001.04} & 1004.04 & 70.91 & \textbf{0.00} \\
p05 & 100 & 2 & 5 & 750.03* & 753.34 & 493.00 & 0.44 & 751.26 & 32.40 & 0.16 & \textbf{750.03} & 17.28 & \textbf{0.00} & \textbf{750.03} & 8.06 & \textbf{0.00} & \textbf{750.03} & 126.86 & \textbf{0.00} & \textbf{750.03} & 750.14 & 45.39 & \textbf{0.00} \\
p06 & 100 & 3 & 6 & 876.50* & 877.84 & 459.00 & 0.15 & 876.70 & 25.11 & 0.02 & \textbf{876.50} & 18.36 & \textbf{0.00} & \textbf{876.50} & 31.93 & \textbf{0.00} & \textbf{876.50} & 175.74 & \textbf{0.00} & \textbf{876.50} & 878.70 & 29.14 & \textbf{0.00} \\
p07 & 100 & 4 & 4 & 881.97* & 891.95 & 463.00 & 1.13 & \textbf{881.97} & 23.76 & \textbf{0.00} & \textbf{881.97} & 25.11 & \textbf{0.00} & 884.66 & 32.86 & 0.30 & \textbf{881.97} & 208.58 & \textbf{0.00} & \textbf{881.97} & 886.33 & 80.74 & \textbf{0.00} \\
p08 & 249 & 2 & 14 & 4369.95 & 4482.44 & 1526.00 & 2.57 & 4390.80 & 89.91 & 0.48 & 4372.78 & 162.00 & 0.06 & 4371.66 & 88.35 & 0.04 & \textbf{4369.95} & 198.87 & \textbf{0.00} & 4372.78 & 4384.34 & 278.04 & 0.06 \\
p09 & 249 & 3 & 12 & 3858.66 & 3920.85 & 1604.00 & 1.61 & 3873.64 & 97.47 & 0.39 & \textbf{3858.66} & 153.90 & \textbf{0.00} & 3880.85 & 79.36 & 0.58 & 3881.31 & 338.69 & 0.59 & \textbf{3858.66} & 3868.83 & 470.09 & \textbf{0.00} \\
p10 & 249 & 4 & 8 & 3629.60 & 3714.65 & 1530.00 & 2.34 & 3650.04 & 98.01 & 0.56 & 3631.11 & 159.03 & 0.04 & \textbf{3629.60} & 82.77 & \textbf{0.00} & 3644.51 & 361.99 & 0.41 & 3631.11 & 3640.81 & 256.83 & 0.04 \\
p11 & 249 & 5 & 6 & 3545.18 & 3580.84 & 1555.00 & 1.01 & 3546.06 & 96.39 & 0.02 & 3546.06 & 115.56 & 0.02 & \textbf{3545.18} & 59.52 & \textbf{0.00} & 3563.28 & 289.53 & 0.51 & 3546.06 & 3548.75 & 276.50 & 0.02 \\
p12 & 80 & 2 & 5 & 1318.95* & \textbf{1318.95} & 334.00 & \textbf{0.00} & \textbf{1318.95} & 20.25 & \textbf{0.00} & \textbf{1318.95} & 8.37 & \textbf{0.00} & \textbf{1318.95} & 1.86 & \textbf{0.00} & \textbf{1318.95} & 74.90 & \textbf{0.00} & \textbf{1318.95} & 1318.95 & 21.40 & \textbf{0.00} \\
p13 & 80 & 2 & 5 & 1318.95* & \textbf{1318.95} & 335.00 & \textbf{0.00} & \textbf{1318.95} & 16.20 & \textbf{0.00} & \textbf{1318.95} & 9.18 & \textbf{0.00} & \textbf{1318.95} & 2.17 & \textbf{0.00} & \textbf{1318.95} & 4.41 & \textbf{0.00} & \textbf{1318.95} & 1318.95 & 23.90 & \textbf{0.00} \\
p14 & 80 & 2 & 5 & 1360.12* & \textbf{1360.12} & 326.00 & \textbf{0.00} & \textbf{1360.12} & 15.66 & \textbf{0.00} & \textbf{1360.12} & 8.91 & \textbf{0.00} & \textbf{1360.12} & 1.86 & \textbf{0.00} & \textbf{1360.12} & 3.16 & \textbf{0.00} & \textbf{1360.12} & 1360.68 & 33.01 & \textbf{0.00} \\
p15 & 160 & 4 & 5 & 2505.42* & 2534.13 & 844.00 & 1.15 & \textbf{2505.42} & 68.31 & \textbf{0.00} & \textbf{2505.42} & 31.05 & \textbf{0.00} & \textbf{2505.42} & 35.34 & \textbf{0.00} & \textbf{2505.42} & 498.14 & \textbf{0.00} & \textbf{2505.42} & 2508.05 & 78.59 & \textbf{0.00} \\
p16 & 160 & 4 & 5 & 2572.23* & \textbf{2572.23} & 843.00 & \textbf{0.00} & \textbf{2572.23} & 50.76 & \textbf{0.00} & \textbf{2572.23} & 31.86 & \textbf{0.00} & \textbf{2572.23} & 36.58 & \textbf{0.00} & \textbf{2572.23} & 20.05 & \textbf{0.00} & \textbf{2572.23} & 2572.23 & 86.40 & \textbf{0.00} \\
p17 & 160 & 4 & 5 & 2709.09* & 2720.23 & 822.00 & 0.41 & \textbf{2709.09} & 48.33 & \textbf{0.00} & \textbf{2709.09} & 34.56 & \textbf{0.00} & \textbf{2709.09} & 33.48 & \textbf{0.00} & 2731.37 & 11.02 & 0.82 & \textbf{2709.09} & 2709.09 & 115.51 & \textbf{0.00} \\
p18 & 240 & 6 & 5 & 3702.85* & 3710.49 & 1491.00 & 0.21 & \textbf{3702.85} & 113.13 & \textbf{0.00} & \textbf{3702.85} & 73.17 & \textbf{0.00} & \textbf{3702.85} & 86.18 & \textbf{0.00} & 3728.49 & 433.42 & 0.69 & \textbf{3702.85} & 3705.41 & 226.93 & \textbf{0.00} \\
p19 & 240 & 6 & 5 & 3827.06* & \textbf{3827.06} & 1512.00 & \textbf{0.00} & \textbf{3827.06} & 85.05 & \textbf{0.00} & \textbf{3827.06} & 68.04 & \textbf{0.00} & \textbf{3827.06} & 79.36 & \textbf{0.00} & \textbf{3827.06} & 25.66 & \textbf{0.00} & \textbf{3827.06} & 3827.06 & 161.36 & \textbf{0.00} \\
p20 & 240 & 6 & 5 & 4058.07* & \textbf{4058.07} & 1483.00 & \textbf{0.00} & \textbf{4058.07} & 81.00 & \textbf{0.00} & \textbf{4058.07} & 70.74 & \textbf{0.00} & \textbf{4058.07} & 82.77 & \textbf{0.00} & 4097.06 & 16.01 & 0.96 & \textbf{4058.07} & 4069.83 & 213.20 & \textbf{0.00} \\
p21 & 360 & 9 & 5 & 5474.84 & 5535.99 & 2890.00 & 1.12 & \textbf{5474.84} & 157.14 & \textbf{0.00} & \textbf{5474.84} & 162.00 & \textbf{0.00} & \textbf{5474.84} & 83.08 & \textbf{0.00} & 5506.26 & 1479.96 & 0.57 & \textbf{5474.84} & 5481.71 & 565.48 & \textbf{0.00} \\
p22 & 360 & 9 & 5 & 5702.16 & 5716.01 & 2934.00 & 0.24 & \textbf{5702.16} & 124.74 & \textbf{0.00} & \textbf{5702.16} & 162.00 & \textbf{0.00} & \textbf{5702.16} & 81.22 & \textbf{0.00} & \textbf{5702.16} & 51.34 & \textbf{0.00} & \textbf{5702.16} & 5702.16 & 289.67 & \textbf{0.00} \\
p23 & 360 & 9 & 5 & 6078.75 & 6139.73 & 2872.00 & 1.00 & \textbf{6078.75} & 119.61 & \textbf{0.00} & \textbf{6078.75} & 162.00 & \textbf{0.00} & 6095.46 & 88.35 & 0.27 & 6145.58 & 30.54 & 1.10 & \textbf{6078.75} & 6114.51 & 666.65 & \textbf{0.00} \\
pr01 & 48 & 4 & 2 & 861.32 & \textbf{861.32} & 242.00 & \textbf{0.00} & \textbf{861.32} & 8.10 & \textbf{0.00} & \textbf{861.32} & 2.70 & \textbf{0.00} & \textbf{861.32} & 1.24 & \textbf{0.00} & \textbf{861.32} & 8.24 & \textbf{0.00} & \textbf{861.32} & 861.32 & 3.36 & \textbf{0.00} \\
pr02 & 96 & 4 & 4 & 1296.25 & 1314.99 & 505.00 & 1.45 & 1307.34 & 27.81 & 0.86 & 1307.34 & 12.42 & 0.86 & 1311.11 & 3.41 & 1.15 & \textbf{1296.25} & 132.71 & \textbf{0.00} & \textbf{1296.25} & 1297.22 & 32.28 & \textbf{0.00} \\
pr03 & 144 & 4 & 6 & 1803.80 & 1815.62 & 854.00 & 0.66 & 1806.60 & 57.78 & 0.16 & \textbf{1803.80} & 31.05 & \textbf{0.00} & \textbf{1803.80} & 36.58 & \textbf{0.00} & \textbf{1803.80} & 100.43 & \textbf{0.00} & \textbf{1803.80} & 1803.88 & 116.61 & \textbf{0.00} \\
pr04 & 192 & 4 & 8 & 2058.31 & 2094.24 & 1158.00 & 1.75 & 2060.93 & 79.92 & 0.13 & 2058.31 & 84.51 & 0.00 & 2064.11 & 38.44 & 0.28 & 2060.74 & 605.31 & 0.12 & \underline{\textbf{2042.45}} & 2054.55 & 246.49 & \textbf{-0.77} \\
pr05 & 240 & 4 & 10 & 2331.20 & 2408.10 & 1529.00 & 3.30 & 2337.84 & 100.44 & 0.28 & 2331.20 & 154.98 & 0.00 & 2349.63 & 66.03 & 0.79 & 2346.72 & 1046.48 & 0.67 & \underline{\textbf{2327.59}} & 2329.24 & 679.93 & \textbf{-0.15} \\
pr06 & 288 & 4 & 12 & 2674.07 & 2768.13 & 2007.00 & 3.52 & 2687.60 & 125.55 & 0.51 & 2676.30 & 162.00 & 0.08 & 2710.30 & 72.54 & 1.35 & 2674.07 & 1275.84 & 0.00 & \underline{\textbf{2663.56}} & 2666.02 & 874.44 & \textbf{-0.39} \\
pr07 & 72 & 6 & 3 & 1082.93 & 1092.12 & 412.00 & 0.85 & 1089.56 & 15.66 & 0.61 & 1089.56 & 5.40 & 0.61 & 1089.56 & 3.41 & 0.61 & 1082.93 & 14.43 & 0.00 & \underline{\textbf{1075.12}} & 1076.60 & 6.77 & \textbf{-0.72} \\
pr08 & 144 & 6 & 6 & 1664.60 & 1676.26 & 906.00 & 0.70 & 1664.85 & 55.89 & 0.02 & 1664.85 & 33.21 & 0.02 & 1665.50 & 20.46 & 0.05 & 1664.60 & 339.80 & 0.00 & \underline{\textbf{1658.23}} & 1662.12 & 233.30 & \textbf{-0.38} \\
pr09 & 216 & 6 & 9 & 2133.20 & 2176.79 & 1462.00 & 2.04 & 2136.42 & 94.50 & 0.15 & 2133.20 & 98.82 & 0.00 & 2151.45 & 48.36 & 0.86 & 2133.20 & 250.60 & 0.00 & \underline{\textbf{2131.70}} & 2139.99 & 212.73 & \textbf{-0.07} \\
pr10 & 288 & 6 & 12 & 2820.88 & 3089.62 & 2105.00 & 9.53 & 2889.82 & 122.85 & 2.44 & 2868.20 & 162.00 & 1.68 & 2910.78 & 93.62 & 3.19 & 2820.88 & 1342.67 & 0.00 & \underline{\textbf{2805.53}} & 2808.77 & 1141.00 & \textbf{-0.54} \\  \hline
Mean & \multicolumn{1}{c}{-} & \multicolumn{1}{c}{-} & \multicolumn{1}{c}{-} & 2423.02 & 2455.56 & 1112.27 & 1.16 & 2428.30 & 63.88 & 0.21 & 2425.22 & 68.66 & 0.10 & 2430.12 & 43.03 & 0.29 & 2430.86 & 293.45 & 0.20 & 2421.33 & 2425.52 & 229.61 & -0.09 \\ \hline
\end{tabular}%
}
\end{table}
\begin{table}[htbp]
\caption{Comparative results on the 10 \emph{C97-T} MDVRP benchmark instances between Optima, SCA and MDFIHA.}
\label{tab:exp-md-c97-t}
\centering
\resizebox{0.8\textwidth}{!}{%
\begin{tabular}{lrrrrrrcrrrr}
\hline
\multirow{2}{*}{Instance} & \multicolumn{1}{c}{\multirow{2}{*}{$N_C$}} & \multicolumn{1}{c}{\multirow{2}{*}{$N_D$}} & \multicolumn{1}{c}{\multirow{2}{*}{$N_V$}} & \multicolumn{1}{c}{\multirow{2}{*}{Optima}} & \multicolumn{3}{c}{SCA} & \multicolumn{4}{c}{MDFIHA (Ours)} \\ \cmidrule(l){6-8} \cmidrule(l){9-12}
 & \multicolumn{1}{c}{} & \multicolumn{1}{c}{} & \multicolumn{1}{c}{} & \multicolumn{1}{c}{} & \multicolumn{1}{c}{$f_{Best}$} & \multicolumn{1}{c}{$\gamma \cdot t$} & \multicolumn{1}{c}{Gap (\%)} & \multicolumn{1}{c}{$f_{Best}$} & \multicolumn{1}{c}{$f_{Avg}$} & \multicolumn{1}{c}{$\gamma \cdot t$} & \multicolumn{1}{c}{Gap (\%)} \\ \hline
pr01 & 48 & 4 & 1 & 861.32 & \textbf{861.32} & 19.58 & \textbf{0.00} & \textbf{861.32} & 861.32 & 4.12 & \textbf{0.00} \\
pr02 & 96 & 4 & 2 & 1307.34 & \textbf{1307.34} & 140.68 & \textbf{0.00} & \textbf{1307.34} & 1308.00 & 28.52 & \textbf{0.00} \\
pr03 & 144 & 4 & 3 & 1803.80 & \textbf{1803.80} & 107.60 & \textbf{0.00} & \textbf{1803.80} & 1804.13 & 95.96 & \textbf{0.00} \\
pr04 & 192 & 4 & 4 & 2058.31 & 2060.93 & 609.62 & 0.13 & \textbf{2058.31} & 2066.65 & 235.54 & \textbf{0.00} \\
pr05 & 240 & 4 & 5 & 2331.20 & 2346.05 & 1038.39 & 0.64 & \textbf{2331.20} & 2339.81 & 797.08 & \textbf{0.00} \\
pr06 & 288 & 4 & 6 & 2676.30 & \textbf{2676.30} & 1288.80 & \textbf{0.00} & \textbf{2676.30} & 2680.60 & 898.26 & \textbf{0.00} \\
pr07 & 72 & 6 & 1 & 1089.56 & \textbf{1089.56} & 10.00 & \textbf{0.00} & \textbf{1089.56} & 1089.56 & 11.35 & \textbf{0.00} \\
pr08 & 144 & 6 & 2 & 1664.85 & \textbf{1664.85} & 339.67 & \textbf{0.00} & \textbf{1664.85} & 1665.52 & 145.92 & \textbf{0.00} \\
pr09 & 216 & 6 & 3 & 2133.20 & \textbf{2133.20} & 240.21 & \textbf{0.00} & \textbf{2133.20} & 2140.07 & 287.91 & \textbf{0.00} \\
pr10 & 288 & 6 & 4 & 2867.26 & 2868.20 & 1344.02 & 0.03 & \textbf{2867.26} & 2875.50 & 1317.23 & \textbf{0.00} \\ \hline
Mean & \multicolumn{1}{c}{-} & \multicolumn{1}{c}{-} & \multicolumn{1}{c}{-} & 1879.31 & 1881.16 & 513.86 & 0.08 & 1879.31 & 1883.12 & 382.19 & 0.00 \\ \hline
\end{tabular}%
}
\end{table}

\begin{table}[]
\caption{Comparative results on the 20 \emph{C01} MDVRPTW benchmark instances between BKS, VCGP, SCA and MDFIHA.}
\label{tab:exp-mdtw-c01}
\resizebox{\textwidth}{!}{%
\begin{tabular}{lrrrrrrcrrcrrrr}
\hline
\multirow{2}{*}{Instance} & \multicolumn{1}{c}{\multirow{2}{*}{$N_C$}} & \multicolumn{1}{c}{\multirow{2}{*}{$N_D$}} & \multicolumn{1}{c}{\multirow{2}{*}{$N_V$}} & \multicolumn{1}{c}{\multirow{2}{*}{BKS}} & \multicolumn{3}{c}{VCGP} & \multicolumn{3}{c}{SCA} & \multicolumn{4}{c}{MDFIHA (Ours)} \\ \cmidrule(l){6-8} \cmidrule(l){9-11} \cmidrule(l){12-15}
 & \multicolumn{1}{c}{} & \multicolumn{1}{c}{} & \multicolumn{1}{c}{} & \multicolumn{1}{c}{} & \multicolumn{1}{c}{$f_{Best}$} & \multicolumn{1}{c}{$\gamma \cdot t$} & \multicolumn{1}{c}{Gap (\%)} & \multicolumn{1}{c}{$f_{Best}$} & \multicolumn{1}{c}{$\gamma \cdot t$} & \multicolumn{1}{c}{Gap (\%)} & \multicolumn{1}{c}{$f_{Best}$} & \multicolumn{1}{c}{$f_{Avg}$} & \multicolumn{1}{c}{$\gamma \cdot t$} & \multicolumn{1}{c}{Gap (\%)} \\ \hline
pr01 & 48 & 4 & 2 & 1074.12 & \textbf{1074.12} & 13.39 & \textbf{0.00} & \textbf{1074.12} & 67.74 & \textbf{0.00} & \textbf{1074.12} & 1074.12 & 14.09 & \textbf{0.00} \\
pr02 & 96 & 4 & 3 & 1762.21 & \textbf{1762.21} & 49.68 & \textbf{0.00} & \textbf{1762.21} & 337.46 & \textbf{0.00} & \textbf{1762.21} & 1762.29 & 91.89 & \textbf{0.00} \\
pr03 & 144 & 4 & 4 & 2373.65 & \textbf{2373.65} & 75.60 & \textbf{0.00} & \textbf{2373.65} & 520.38 & \textbf{0.00} & \textbf{2373.65} & 2380.24 & 166.76 & \textbf{0.00} \\
pr04 & 192 & 4 & 5 & 2814.34 & 2815.75 & 254.45 & 0.05 & \textbf{2814.34} & 3454.26 & \textbf{0.00} & \textbf{2814.34} & 2817.89 & 669.76 & \textbf{0.00} \\
pr05 & 240 & 4 & 6 & 2964.65 & 2964.65 & 374.98 & 0.00 & 2965.18 & 915.85 & 0.02 & \underline{\textbf{2962.25}} & 2967.52 & 1494.27 & \textbf{-0.08} \\
pr06 & 288 & 4 & 7 & 3588.78 & \textbf{3588.78} & 580.18 & \textbf{0.00} & 3590.58 & 4026.62 & 0.05 & \textbf{3588.78} & 3594.42 & 1516.47 & \textbf{0.00} \\
pr07 & 72 & 6 & 2 & 1418.22 & \textbf{1418.22} & 22.03 & \textbf{0.00} & \textbf{1418.22} & 146.58 & \textbf{0.00} & \textbf{1418.22} & 1418.93 & 56.07 & \textbf{0.00} \\
pr08 & 144 & 6 & 3 & 2096.73 & \textbf{2096.73} & 103.25 & \textbf{0.00} & \textbf{2096.73} & 501.21 & \textbf{0.00} & \textbf{2096.73} & 2100.42 & 239.77 & \textbf{0.00} \\
pr09 & 216 & 6 & 4 & 2712.56 & \textbf{2712.56} & 224.64 & \textbf{0.00} & 2717.69 & 1539.15 & 0.19 & \textbf{2712.56} & 2713.54 & 437.81 & \textbf{0.00} \\
pr10 & 288 & 6 & 5 & 3465.92 & 3465.92 & 657.50 & 0.00 & 3469.29 & 4658.51 & 0.10 & \underline{\textbf{3465.54}} & 3477.00 & 1334.49 & \textbf{-0.01} \\
pr11 & 48 & 4 & 1 & 1005.73 & \textbf{1005.73} & 22.03 & \textbf{0.00} & \textbf{1005.73} & 13.82 & \textbf{0.00} & \textbf{1005.73} & 1005.73 & 34.08 & \textbf{0.00} \\
pr12 & 96 & 4 & 2 & 1464.50 & \textbf{1464.50} & 72.58 & \textbf{0.00} & \textbf{1464.50} & 480.07 & \textbf{0.00} & \textbf{1464.50} & 1465.29 & 104.60 & \textbf{0.00} \\
pr13 & 144 & 4 & 3 & 2001.81 & \textbf{2001.81} & 127.01 & \textbf{0.00} & \textbf{2001.81} & 314.08 & \textbf{0.00} & \textbf{2001.81} & 2001.82 & 170.77 & \textbf{0.00} \\
pr14 & 192 & 4 & 4 & 2195.33 & \textbf{2195.33} & 282.96 & \textbf{0.00} & \textbf{2195.33} & 1924.17 & \textbf{0.00} & \textbf{2195.33} & 2197.15 & 248.47 & \textbf{0.00} \\
pr15 & 240 & 4 & 5 & 2433.15 & \textbf{2433.15} & 542.59 & \textbf{0.00} & 2434.94 & 1583.78 & 0.07 & \textbf{2433.15} & 2449.23 & 1685.11 & \textbf{0.00} \\ 			
pr16 & 288 & 4 & 6 & 2836.67 & \textbf{2836.67} & 689.90 & \textbf{0.00} & 2850.69 & 2521.00 & 0.49 & 2837.85 & 2842.35 & 2877.91 & 0.04 \\
pr17 & 72 & 6 & 1 & 1236.24 & \textbf{1236.24} & 45.36 & \textbf{0.00} & \textbf{1236.24} & 79.47 & \textbf{0.00} & \textbf{1236.24} & 1236.62 & 68.23 & \textbf{0.00} \\
pr18 & 144 & 6 & 2 & 1788.18 & \textbf{1788.18} & 142.56 & \textbf{0.00} & \textbf{1788.18} & 378.07 & \textbf{0.00} & \textbf{1788.18} & 1788.71 & 197.17 & \textbf{0.00} \\
pr19 & 216 & 6 & 3 & 2261.08 & 2261.08 & 371.09 & 0.00 & 2263.74 & 1304.18 & 0.12 & \underline{\textbf{2257.13}} & 2266.69 & 552.22 & \textbf{-0.17} \\
pr20 & 288 & 6 & 4 & 2993.31 & 2993.31 & 958.18 & 0.00 & 2995.08 & 2850.71 & 0.06 & \underline{\textbf{2983.78}} & 2996.78 & 2694.24 & \textbf{-0.32} \\ \hline
Mean & \multicolumn{1}{c}{-} & \multicolumn{1}{c}{-} & \multicolumn{1}{c}{-} & 2224.42 & 2224.43 & 280.50 & 0.00 & 2225.91 & 1380.86 & 0.05 & 2223.61 & 2227.84 & 732.71 & -0.03 \\ \hline
\end{tabular}%
}
\end{table}
\begin{table}[]
\caption{Comparative results on the 10 \emph{C01-R} MDVRPTW benchmark instances between BKS, VCGP and MDFIHA.}
\label{tab:exp-mdtw-c01-r}
\centering
\resizebox{0.8\textwidth}{!}{%
\begin{tabular}{lrrrrrrcrrrr}
\hline
\multirow{2}{*}{Instance} & \multicolumn{1}{c}{\multirow{2}{*}{$N_C$}} & \multicolumn{1}{c}{\multirow{2}{*}{$N_D$}} & \multicolumn{1}{c}{\multirow{2}{*}{$N_V$}} & \multicolumn{1}{c}{\multirow{2}{*}{BKS}} & \multicolumn{3}{c}{VCGP} & \multicolumn{4}{c}{MDFIHA (Ours)} \\ \cmidrule(l){6-8} \cmidrule(l){9-12}
 & \multicolumn{1}{c}{} & \multicolumn{1}{c}{} & \multicolumn{1}{c}{} & \multicolumn{1}{c}{} & \multicolumn{1}{c}{$f_{Best}$} & \multicolumn{1}{c}{$\gamma \cdot t$} & \multicolumn{1}{c}{Gap (\%)} & \multicolumn{1}{c}{$f_{Best}$} & \multicolumn{1}{c}{$f_{Avg}$} & \multicolumn{1}{c}{$\gamma \cdot t$} & \multicolumn{1}{c}{Gap (\%)} \\ \hline
pr11 & 48 & 4 & 2 & 1005.73 & 1005.73 & 22.03 & 0.00 & \underline{\textbf{922.44}} & 922.44 & 9.05 & \textbf{-8.28} \\
pr12 & 96 & 4 & 3 & 1464.50 & 1464.50 & 72.58 & 0.00 & \underline{\textbf{1430.29}} & 1430.29 & 55.31 & \textbf{-2.34} \\
pr13 & 144 & 4 & 4 & 2001.81 & \textbf{2001.81} & 127.01 & \textbf{0.00} & \textbf{2001.81} & 2002.33 & 175.36 & \textbf{0.00} \\
pr14 & 192 & 4 & 5 & 2195.33 & 2195.33 & 282.96 & 0.00 & \underline{\textbf{2191.06}} & 2192.77 & 285.07 & \textbf{-0.19} \\
pr15 & 240 & 4 & 6 & 2433.15 & 2433.15 & 542.59 & 0.00 & \underline{\textbf{2419.88}} & 2425.73 & 900.55 & \textbf{-0.55} \\
pr16 & 288 & 4 & 7 & 2836.67 & 2836.67 & 689.90 & 0.00 & \underline{\textbf{2823.68}} & 2833.65 & 1432.80 & \textbf{-0.46} \\
pr17 & 72 & 6 & 2 & 1236.24 & 1236.24 & 45.36 & 0.00 & \underline{\textbf{1167.20}} & 1167.20 & 32.11 & \textbf{-5.58} \\
pr18 & 144 & 6 & 3 & 1788.18 & 1788.18 & 142.56 & 0.00 & \underline{\textbf{1775.25}} & 1780.84 & 242.02 & \textbf{-0.72} \\
pr19 & 216 & 6 & 4 & 2261.08 & 2261.08 & 371.09 & 0.00 & \underline{\textbf{2256.11}} & 2266.39 & 1103.20 & \textbf{-0.22} \\
pr20 & 288 & 6 & 5 & 2993.31 & 2993.31 & 958.18 & 0.00 & \underline{\textbf{2900.86}} & 2903.34 & 1307.47 & \textbf{-3.09} \\ \hline
Mean & \multicolumn{1}{c}{-} & \multicolumn{1}{c}{-} & \multicolumn{1}{c}{-} & 2021.60 & 2021.60 & 325.43 & 0.00 & 1988.86 & 1992.50 & 554.29 & -2.14 \\ \hline
\end{tabular}%
}
\end{table}
\begin{table}[]
\caption{Comparative results on the 28 \emph{V13} MDVRPTW benchmark instances between VCGP, MDFIHA and MDFIHA-ETGA.}
\label{tab:exp-mdtw-v13}
\resizebox{\textwidth}{!}{%
\begin{tabular}{lrrrrrcrrrcrrrr}
\hline
\multirow{2}{*}{Instance} & \multicolumn{1}{c}{\multirow{2}{*}{$N_C$}} & \multicolumn{1}{c}{\multirow{2}{*}{$N_D$}} & \multicolumn{1}{c}{\multirow{2}{*}{$N_V$}} & \multicolumn{3}{c}{VCGP} & \multicolumn{4}{c}{MDFIHA (Ours)} & \multicolumn{4}{c}{MDFIHA-ETGA (Ours)} \\ \cmidrule(l){5-7} \cmidrule(l){8-11} \cmidrule(l){12-15}
 & \multicolumn{1}{c}{} & \multicolumn{1}{c}{} & \multicolumn{1}{c}{} & \multicolumn{1}{c}{$f_{Best}$} & \multicolumn{1}{c}{$f_{Avg}$} & \multicolumn{1}{c}{$\gamma \cdot t$} & \multicolumn{1}{c}{$f_{Best}$} & \multicolumn{1}{c}{$f_{Avg}$} & \multicolumn{1}{c}{$\gamma \cdot t$} & \multicolumn{1}{c}{Gap (\%)} & \multicolumn{1}{c}{$f_{Best}$} & \multicolumn{1}{c}{$f_{Avg}$} & \multicolumn{1}{c}{$\gamma \cdot t$} & \multicolumn{1}{c}{Gap (\%)} \\ \hline
pr11a & 360 & 4 & 10 & 6720.71 & 6772.00 & 726.19 & \underline{\textbf{6685.36}} & 6702.83 & 2868.55 & \textbf{-0.53} & \underline{\textbf{6683.00}} & 6697.60 & 5167.52 & \textbf{-0.56} \\
pr12a & 480 & 4 & 13 & 8179.80 & 8259.57 & 1296.00 & 8208.50 & 8225.16 & 7099.69 & 0.35 & \underline{\textbf{8164.72}} & 8191.44 & 6226.49 & \textbf{-0.18} \\
pr13a & 600 & 4 & 16 & 9667.20 & 9751.22 & 2369.52 & \underline{\textbf{9602.32}} & 9662.21 & 7201.87 & \textbf{-0.67} & \underline{\textbf{9571.71}} & 9616.67 & 7015.26 & \textbf{-0.99} \\
pr14a & 720 & 4 & 19 & 11124.01 & 11235.13 & 2836.08 & \underline{\textbf{11089.24}} & 11117.63 & 7202.28 & \textbf{-0.31} & \underline{\textbf{11062.70}} & 11110.91 & 6919.64 & \textbf{-0.55} \\
pr15a & 840 & 4 & 22 & 13013.97 & 13078.26 & 5721.41 & 13019.83 & 13080.84 & 7205.23 & 0.05 & \underline{\textbf{12895.63}} & 12941.05 & 7201.95 & \textbf{-0.91} \\
pr16a & 960 & 4 & 26 & 14299.87 & 14415.89 & 5772.82 & 14450.22 & 14489.24 & 7208.66 & 1.05 & \underline{\textbf{14223.38}} & 14319.36 & 7089.71 & \textbf{-0.53} \\
pr17a & 360 & 6 & 7 & 6304.30 & 6340.66 & 744.34 & 6309.20 & 6321.09 & 3615.48 & 0.08 & 6314.89 & 6335.83 & 4280.81 & 0.17 \\
pr18a & 520 & 6 & 10 & 8308.32 & 8381.71 & 1911.60 & \underline{\textbf{8228.31}} & 8247.81 & 7202.66 & \textbf{-0.96} & \underline{\textbf{8238.25}} & 8267.51 & 6809.99 & \textbf{-0.84} \\
pr19a & 700 & 6 & 13 & 10677.61 & 10734.60 & 3214.94 & 10679.74 & 10728.21 & 7203.97 & 0.02 & \underline{\textbf{10663.75}} & 10675.57 & 6429.51 & \textbf{-0.13} \\
pr20a & 880 & 6 & 16 & 11963.91 & 12142.60 & 4638.38 & \underline{\textbf{11929.23}} & 11983.04 & 7208.83 & \textbf{-0.29} & \underline{\textbf{11839.92}} & 11864.35 & 7201.67 & \textbf{-1.04} \\
pr21a & 420 & 12 & 4 & 6260.53 & 6321.20 & 1209.60 & 6261.83 & 6275.48 & 4780.65 & 0.02 & \underline{\textbf{6258.84}} & 6271.13 & 5903.00 & \textbf{-0.03} \\
pr22a & 600 & 12 & 6 & 7985.37 & 8047.87 & 3285.36 & \underline{\textbf{7919.77}} & 7955.53 & 7203.30 & \textbf{-0.82} & \underline{\textbf{7935.78}} & 7965.98 & 6742.34 & \textbf{-0.62} \\
pr23a & 780 & 12 & 8 & 9937.43 & 9984.75 & 5949.50 & \underline{\textbf{9893.39}} & 9960.04 & 7204.03 & \textbf{-0.44} & \underline{\textbf{9870.23}} & 9891.89 & 7203.07 & \textbf{-0.68} \\
pr24a & 960 & 12 & 10 & 11923.72 & 11971.74 & 8517.74 & 12031.09 & 12063.55 & 7210.03 & 0.90 & \underline{\textbf{11861.22}} & 11882.24 & 7204.22 & \textbf{-0.52} \\
pr11b & 360 & 4 & 8 & 4839.44 & 4852.67 & 779.33 & \underline{\textbf{4832.98}} & 4842.14 & 3257.81 & \textbf{-0.13} & \underline{\textbf{4831.74}} & 4842.88 & 5057.87 & \textbf{-0.16} \\
pr12b & 480 & 4 & 11 & 6063.26 & 6084.33 & 1256.69 & \underline{\textbf{6038.38}} & 6053.57 & 7018.54 & \textbf{-0.41} & \underline{\textbf{6033.93}} & 6063.03 & 6157.79 & \textbf{-0.48} \\
pr13b & 600 & 4 & 14 & 7254.17 & 7282.25 & 3066.77 & \underline{\textbf{7245.29}} & 7268.61 & 7204.84 & \textbf{-0.12} & \underline{\textbf{7231.95}} & 7274.66 & 6866.29 & \textbf{-0.31} \\
pr14b & 720 & 4 & 17 & 8732.29 & 8796.77 & 4273.34 & 8772.44 & 8809.61 & 7203.77 & 0.46 & 8761.07 & 8783.31 & 7200.69 & 0.33 \\
pr15b & 840 & 4 & 20 & 10439.72 & 10496.39 & 5593.54 & \underline{\textbf{10425.82}} & 10488.71 & 7204.47 & \textbf{-0.13} & \underline{\textbf{10417.15}} & 10441.08 & 6845.19 & \textbf{-0.22} \\
pr16b & 960 & 4 & 23 & 11483.22 & 11565.39 & 7357.39 & 11568.44 & 11595.45 & 7210.55 & 0.74 & \underline{\textbf{11394.50}} & 11457.72 & 7204.34 & \textbf{-0.77} \\
pr17b & 360 & 6 & 6 & 4806.01 & 4847.58 & 681.70 & \underline{\textbf{4790.27}} & 4809.84 & 2545.24 & \textbf{-0.33} & \underline{\textbf{4797.43}} & 4811.54 & 4601.24 & \textbf{-0.18} \\
pr18b & 520 & 6 & 9 & 6526.72 & 6555.95 & 1704.24 & \underline{\textbf{6467.78}} & 6498.29 & 6939.66 & \textbf{-0.90} & \underline{\textbf{6484.49}} & 6505.49 & 5536.92 & \textbf{-0.65} \\
pr19b & 700 & 6 & 12 & 8227.25 & 8295.30 & 3479.76 & \underline{\textbf{8183.28}} & 8214.23 & 7202.20 & \textbf{-0.53} & \underline{\textbf{8164.09}} & 8176.36 & 6999.77 & \textbf{-0.77} \\
pr20b & 880 & 6 & 15 & 10325.80 & 10378.54 & 6511.97 & \underline{\textbf{10274.63}} & 10305.06 & 7206.93 & \textbf{-0.50} & \underline{\textbf{10162.84}} & 10196.89 & 7202.17 & \textbf{-1.58} \\
pr21b & 420 & 12 & 4 & 4866.57 & 4887.57 & 1587.60 & \underline{\textbf{4830.90}} & 4882.99 & 3005.66 & \textbf{-0.73} & \underline{\textbf{4834.49}} & 4847.04 & 5767.18 & \textbf{-0.66} \\
pr22b & 600 & 12 & 6 & 6488.50 & 6537.15 & 3165.70 & \underline{\textbf{6474.00}} & 6480.90 & 7039.33 & \textbf{-0.22} & \underline{\textbf{6446.58}} & 6479.22 & 6494.99 & \textbf{-0.65} \\
pr23b & 780 & 12 & 7 & 8523.41 & 8603.85 & 7084.37 & \underline{\textbf{8491.96}} & 8516.51 & 7203.59 & \textbf{-0.37} & \underline{\textbf{8417.24}} & 8453.39 & 7201.55 & \textbf{-1.25} \\
pr24b & 960 & 12 & 8 & 10890.08 & 10997.66 & 12895.63 & \underline{\textbf{10721.40}} & 10806.23 & 7203.38 & \textbf{-1.55} & \underline{\textbf{10665.16}} & 10667.91 & 7201.68 & \textbf{-2.07} \\ \hline
Mean & \multicolumn{1}{c}{-} & \multicolumn{1}{c}{-} & \multicolumn{1}{c}{-} & 8779.76 & 8843.52 & 3843.98 & 8765.20 & 8799.46 & 6352.19 & -0.22 & 8722.38 & 8751.15 & 6490.46 & -0.60 \\ \hline
\end{tabular}%
}
\end{table}
\begin{table}[]
\caption{Comparative results on the 24 \emph{L14} MDOVRP benchmark instances between BKS, JB, RM, SCA and MDFIHA.}
\label{tab:exp-mdo-l14}
\resizebox{\textwidth}{!}{%
\begin{tabular}{lrrrrrcrrcrrcrrrr}
\hline
\multirow{2}{*}{Instance} & \multicolumn{1}{c}{\multirow{2}{*}{$N_C$}} & \multicolumn{1}{c}{\multirow{2}{*}{$N_D$}} & \multicolumn{1}{c}{\multirow{2}{*}{BKS}} & \multicolumn{3}{c}{JB} & \multicolumn{3}{c}{RM} & \multicolumn{3}{c}{SCA} & \multicolumn{4}{c}{MDFIHA (Ours)} \\ \cmidrule(l){5-7} \cmidrule(l){8-10} \cmidrule(l){11-13} \cmidrule(l){14-17}
 & \multicolumn{1}{c}{} & \multicolumn{1}{c}{} & \multicolumn{1}{c}{} & \multicolumn{1}{c}{$f_{Best}$} & \multicolumn{1}{c}{$\gamma \cdot t$} & \multicolumn{1}{c}{Gap (\%)} & \multicolumn{1}{c}{$f_{Best}$} & \multicolumn{1}{c}{$\gamma \cdot t$} & \multicolumn{1}{c}{Gap (\%)} & \multicolumn{1}{c}{$f_{Best}$} & \multicolumn{1}{c}{$\gamma \cdot t$} & \multicolumn{1}{c}{Gap (\%)} & \multicolumn{1}{c}{$f_{Best}$} & \multicolumn{1}{c}{$f_{Avg}$} & \multicolumn{1}{c}{$\gamma \cdot t$} & \multicolumn{1}{c}{Gap (\%)} \\ \hline
p01 & 50 & 4 & 386.18* & \textbf{386.18} & 6.84 & \textbf{0.00} & \textbf{386.18} & 0.92 & \textbf{0.00} & \textbf{386.18} & 22.01 & \textbf{0.00} & \textbf{386.18} & 386.91 & 3.83 & \textbf{0.00} \\
p02 & 50 & 4 & 375.93* & 376.44 & 7.83 & 0.14 & \textbf{375.93} & 0.20 & \textbf{0.00} & \textbf{375.93} & 29.15 & \textbf{0.00} & \textbf{375.93} & 375.93 & 2.60 & \textbf{0.00} \\
p03 & 75 & 5 & 474.57* & 475.79 & 14.58 & 0.26 & \textbf{474.57} & 1.39 & \textbf{0.00} & \textbf{474.57} & 53.74 & \textbf{0.00} & \textbf{474.57} & 474.57 & 13.65 & \textbf{0.00} \\
p04 & 100 & 2 & 662.22 & \textbf{662.22} & 27.54 & \textbf{0.00} & \textbf{662.22} & 7272.00 & \textbf{0.00} & \textbf{662.22} & 21.26 & \textbf{0.00} & \textbf{662.22} & 662.22 & 19.46 & \textbf{0.00} \\
p05 & 100 & 2 & 607.53* & 609.04 & 19.98 & 0.25 & \textbf{607.53} & 73.68 & \textbf{0.00} & \textbf{607.53} & 41.67 & \textbf{0.00} & \textbf{607.53} & 608.39 & 21.90 & \textbf{0.00} \\
p06 & 100 & 3 & 611.99* & \textbf{611.99} & 19.80 & \textbf{0.00} & \textbf{611.99} & 416.08 & \textbf{0.00} & \textbf{611.99} & 31.48 & \textbf{0.00} & \textbf{611.99} & 611.99 & 17.19 & \textbf{0.00} \\
p07 & 100 & 4 & 608.28* & 609.60 & 27.54 & 0.22 & \textbf{608.28} & 1041.73 & \textbf{0.00} & \textbf{608.28} & 38.68 & \textbf{0.00} & \textbf{608.28} & 608.77 & 29.11 & \textbf{0.00} \\
p08 & 249 & 2 & 2776.12 & 2794.10 & 133.29 & 0.65 & 2870.21 & 7272.00 & 3.39 & \textbf{2776.12} & 24.25 & \textbf{0.00} & 2776.35 & 2781.67 & 320.77 & 0.01 \\
p09 & 249 & 3 & 2578.49 & 2582.65 & 136.44 & 0.16 & 2660.45 & 7272.00 & 3.18 & 2587.04 & 132.65 & 0.33 & \underline{\textbf{2576.17}} & 2581.95 & 212.03 & \textbf{-0.09} \\
p10 & 249 & 4 & 2482.32 & 2500.43 & 103.77 & 0.73 & 2528.98 & 7272.00 & 1.88 & 2482.32 & 253.03 & 0.00 & \underline{\textbf{2476.12}} & 2479.58 & 167.36 & \textbf{-0.25} \\
p11 & 249 & 5 & 2468.45 & 2478.31 & 119.97 & 0.40 & 2499.25 & 7272.00 & 1.25 & 2472.85 & 338.26 & 0.18 & \underline{\textbf{2454.01}} & 2458.01 & 166.29 & \textbf{-0.58} \\
p12 & 80 & 2 & 953.26* & \textbf{953.26} & 9.36 & \textbf{0.00} & \textbf{953.26} & 0.46 & \textbf{0.00} & \textbf{953.26} & 56.77 & \textbf{0.00} & \textbf{953.26} & 953.26 & 6.41 & \textbf{0.00} \\
p15 & 160 & 4 & 1885.81* & \textbf{1885.81} & 32.94 & \textbf{0.00} & \textbf{1885.81} & 3.67 & \textbf{0.00} & \textbf{1885.81} & 213.04 & \textbf{0.00} & \textbf{1885.81} & 1885.81 & 25.28 & \textbf{0.00} \\
p18 & 240 & 6 & 2818.36* & \textbf{2818.36} & 68.40 & \textbf{0.00} & \textbf{2818.36} & 14.83 & \textbf{0.00} & \textbf{2818.36} & 482.29 & \textbf{0.00} & \textbf{2818.36} & 2818.36 & 61.62 & \textbf{0.00} \\
pr01 & 48 & 4 & 647.03* & \textbf{647.03} & 4.50 & \textbf{0.00} & \textbf{647.03} & 0.14 & \textbf{0.00} & \textbf{647.03} & 57.36 & \textbf{0.00} & \textbf{647.03} & 647.03 & 2.48 & \textbf{0.00} \\
pr02 & 96 & 4 & 979.82* & \textbf{979.82} & 17.19 & \textbf{0.00} & \textbf{979.82} & 1.99 & \textbf{0.00} & \textbf{979.82} & 114.72 & \textbf{0.00} & \textbf{979.82} & 980.33 & 9.66 & \textbf{0.00} \\
pr03 & 144 & 4 & 1423.48* & 1429.38 & 37.98 & 0.41 & \textbf{1423.48} & 17.01 & \textbf{0.00} & \textbf{1423.48} & 203.78 & \textbf{0.00} & \textbf{1423.48} & 1423.82 & 24.56 & \textbf{0.00} \\
pr04 & 192 & 4 & 1514.07* & 1524.18 & 54.45 & 0.67 & \textbf{1514.07} & 1755.56 & \textbf{0.00} & \textbf{1514.07} & 278.91 & \textbf{0.00} & \textbf{1514.07} & 1514.96 & 64.50 & \textbf{0.00} \\
pr05 & 240 & 4 & 1699.40 & 1700.93 & 116.91 & 0.09 & 1716.02 & 7272.00 & 0.98 & 1699.40 & 376.49 & 0.00 & \underline{\textbf{1692.61}} & 1695.94 & 189.34 & \textbf{-0.40} \\
pr06 & 288 & 4 & 1978.46 & 1992.26 & 161.37 & 0.70 & 1978.46 & 7272.00 & 0.00 & 1982.13 & 417.65 & 0.19 & \underline{\textbf{1977.30}} & 1979.97 & 191.71 & \textbf{-0.06} \\
pr07 & 72 & 6 & 821.25* & \textbf{821.25} & 10.62 & \textbf{0.00} & \textbf{821.25} & 0.30 & \textbf{0.00} & \textbf{821.25} & 130.64 & \textbf{0.00} & \textbf{821.25} & 821.25 & 5.10 & \textbf{0.00} \\
pr08 & 144 & 6 & 1254.45* & 1258.64 & 29.16 & 0.33 & \textbf{1254.45} & 8.53 & \textbf{0.00} & \textbf{1254.45} & 252.46 & \textbf{0.00} & \textbf{1254.45} & 1255.83 & 32.68 & \textbf{0.00} \\
pr09 & 216 & 6 & 1591.78* & 1592.77 & 77.13 & 0.06 & \textbf{1591.78} & 329.48 & \textbf{0.00} & \textbf{1591.78} & 414.37 & \textbf{0.00} & \textbf{1591.78} & 1591.78 & 112.89 & \textbf{0.00} \\
pr10 & 288 & 6 & 1969.35 & 1975.36 & 157.05 & 0.31 & 1997.96 & 7272.00 & 1.45 & 1969.35 & 597.89 & 0.00 & \underline{\textbf{1968.67}} & 1972.83 & 195.31 & \textbf{-0.03} \\ \hline
Mean & \multicolumn{1}{c}{-} & \multicolumn{1}{c}{-} & 1398.69 & 1402.74 & 58.11 & 0.22 & 1411.14 & 2576.75 & 0.51 & 1399.38 & 190.94 & 0.03 & 1397.39 & 1398.80 & 78.99 & -0.06 \\ \hline
\end{tabular}%
}
\end{table}

\end{document}